\documentclass[runningheads]{llncs}

\PassOptionsToPackage{table}{xcolor}

\usepackage[final,year=2026,ID=*****]{eccv}

\usepackage{eccvabbrv}

\usepackage{graphicx}
\usepackage{booktabs}
\usepackage{multirow}
\usepackage{placeins}
\usepackage{amsmath}
\usepackage[normalem]{ulem}
\usepackage[breakable,skins]{tcolorbox}

\usepackage[accsupp]{axessibility}

\usepackage[breaklinks,colorlinks,citecolor=eccvblue]{hyperref}

\usepackage{geometry}
\usepackage{caption}
\makeatletter
\renewcommand\@fnsymbol[1]{\ensuremath{\ifcase#1\or *\or \dagger\or \ddagger\or
  \mathsection\or \mathparagraph\fi}}
\makeatother

\definecolor{revRed}   {HTML}{D7263D}
\definecolor{revOrange}{HTML}{F46036}
\definecolor{revPurple}{HTML}{8338EC}
\definecolor{revGray}  {HTML}{6C757D}
\definecolor{ourRow}   {HTML}{E8F2FB}

\definecolor{revBlue}{HTML}{1565C0}

\begin{document}

\title{\LARGE CtrlVTON: Controllable Virtual Try-On via Visual-Instance-Prompt Segmentation}
\titlerunning{CtrlVTON}
\authorrunning{S. Lee et al.}
\author{Seungyong Lee\inst{1}\textsuperscript{\textdagger} \and
        Hyun Jun Jang\inst{1} \and
        Sangoh Kim\inst{1,2}\textsuperscript{\ddag} \and
        Sungjoon Park\inst{1}\textsuperscript{\textdagger,*}}
\institute{NXN Labs \and KAIST \\
\email{seungyong@nxn.ai, hjang@nxn.ai, tkddh1109@kaist.ac.kr, sungjoon@nxn.ai} \\
\url{https://github.com/nxnai/CtrlVTON}
}
\maketitle

\makeatletter
\renewcommand{\thefootnote}{}
\footnotetext{\textdagger\ Equal contribution.}
\footnotetext{\ddag\ Work done during internship at NXN Labs.}
\footnotetext{$*$\ Corresponding author.}
\makeatother

\setlength{\parskip}{3pt plus 1pt minus 1pt}

\begin{center}
  \includegraphics[width=\textwidth]{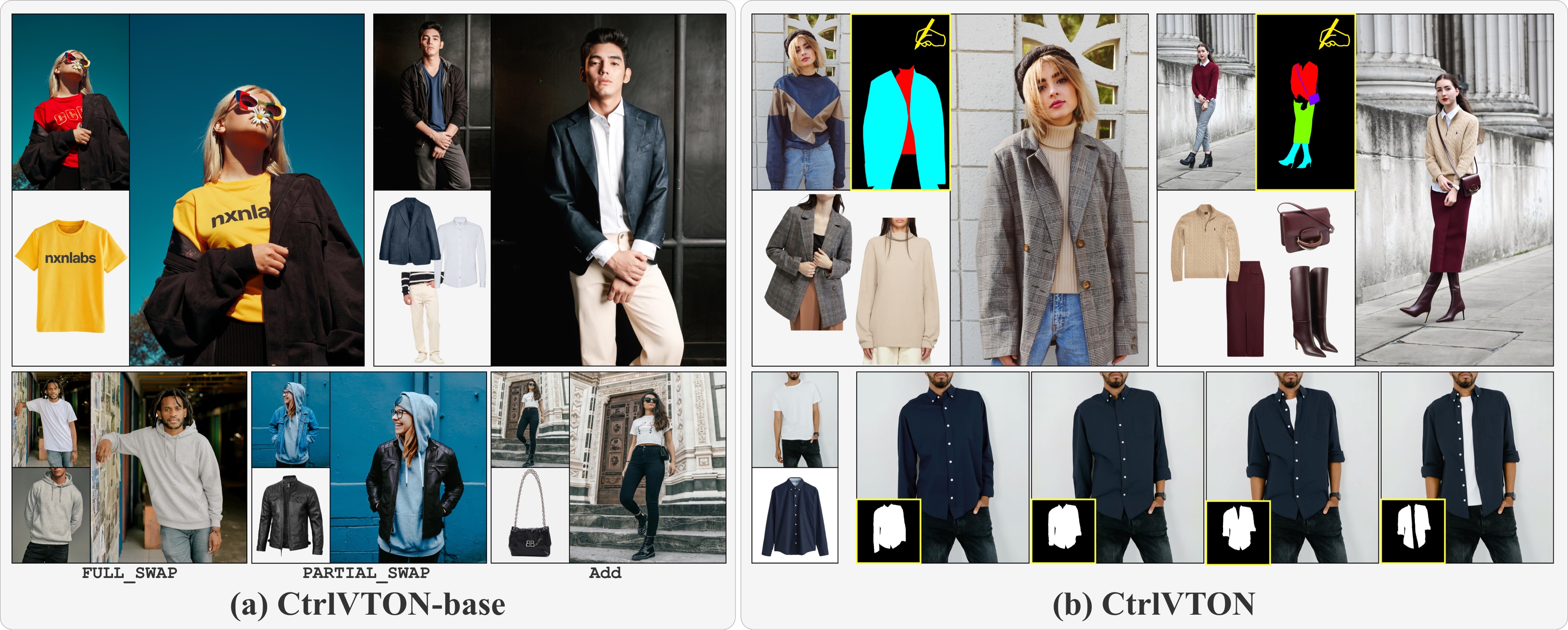}
  \vspace{0pt}
    \captionof{figure}{
    \textbf{(a) CtrlVTON-base} is a baseline image-editing model that enables semantic control via task tokens (\textsc{full\_swap} / \textsc{partial\_swap} / \textsc{add}) over multiple garment classes.
    \textbf{(b) CtrlVTON} enables fine-grained spatial control through hand-drawn masks (\textcolor[HTML]{E8A000}{yellow}), supporting both single- and multi-garment try-on.}
  \label{fig:teaser}
\end{center}
\vspace{-2em}
\begin{abstract}
Virtual try-on (VTO) has made significant progress in realistically transferring garments onto a target person.
Yet most systems give the user little control over how a garment should be worn---its \textbf{size} (loose or fitted), \textbf{style} (\eg, tucked in or untucked, open or closed), and \textbf{spatial placement} on the body.
We address this gap with two complementary contributions.
First, we define and solve \emph{Visual-Instance-Prompt Segmentation} via \textbf{VIP-SAM}: given a flatlay image of a garment, segment that specific instance in a photograph of a person wearing it.
This is an instance-level task, distinct from the typically studied category-level segmentation.
Second, we introduce \textbf{CtrlVTON}, a controllable VTO framework that recasts try-on as an image \emph{editing} problem and adds segmentation masks as pixel-level control over garment layout, including style, size, and spatial placement on the body.
VIP-SAM and CtrlVTON each achieve state-of-the-art results on their respective tasks.
In particular, CtrlVTON generates images that follow user-provided layouts far more faithfully than the strongest proprietary editing systems while matching them on garment fidelity.
\keywords{Virtual Try-On \and Controllable Image Generation \and Visual-Prompt Segmentation}
\end{abstract}

\section{Introduction}
\label{sec:intro}

The fashion and e-commerce industries have long sought to bridge the gap between how a garment appears online and how it looks when worn.
Virtual try-on (VTO) addresses this need by synthesizing a photorealistic image of a person wearing the garment, allowing customers to visualize how they look without physically trying it on.
Recent diffusion-based methods have substantially improved photorealism and garment fidelity~\cite{zhu2023tryondiffusion,morelli2023ladivton,kim2024stableviton,choi2024idmvton,chong2024catvton,lee2025voost}, making VTO commercially viable.
Despite this progress, current VTO methods share a fundamental limitation: they allow users limited control over \emph{how} a garment should be worn, including \textbf{size} (e.g. loose or fitted), \textbf{style} (e.g. tucked in or untucked, zipped or unzipped), and \textbf{spatial placement} (e.g. spatial position, layering).

To enable controllability, we start by recasting VTO as an \emph{image-editing} problem rather than an inpainting problem.
This reformulation avoids the well-known failure modes of inpainting-based VTO methods (Sec.~\ref{sec:supp_inpainting_failures} of Supp.).
The resulting model, \textbf{CtrlVTON-base}, handles diverse garment categories (tops, bottoms, dresses, shoes, bags) and garment display formats (flatlay, on-person, in-the-wild).
It also supports two scenarios that conventional VTO methods struggle with: \emph{garment layering} (adding an item over the outfit) and \emph{selective garment switching} (replacing only a specific item).
We then notice that even in the editing framework, segmentation masks can be used a pixel-level interface for spatial control.
By extending CtrlVTON-base with this capability, we obtain \textbf{CtrlVTON} (Sec.~\ref{sec:cvton}), which not only matches the strongest VTO systems on image quality, but also enables fine-grained control over garment style, size, and placement via segmentation masks.

The models rely critically on our data preparation pipeline, which requires an automatic, scalable way to segment each reference garment within the person image.
We formalize this segmentation problem as a new task, \emph{visual-instance-prompt segmentation} (VIP-Seg): given a support image (\eg, a flatlay garment), locate \emph{the same instance} in a query image (\eg, a person wearing it).
This task is distinct from the category-level visual-reference-prompt segmentation (VRP-Seg) studied in VRP-SAM~\cite{sun2024vrpsam} and related works~\cite{wang2023seggpt,wang2023images,wang2025prosam,liu2023matcher}.
The model must identify a specific instance under same-category distractors, heavy occlusion, and non-rigid deformation between studio flatlay and on-body imagery.
Our model, \textbf{VIP-SAM}, achieves state-of-the-art performance on both our purpose-built fashion benchmark and standard category-level benchmarks~\cite{shaban2017oneshot,nguyen2019fwb} repurposed for instance-level segmentation.

In summary, we make the following contributions:
\begin{enumerate}
  \item We introduce the \textbf{VIP-Seg} task. Our model, \textbf{VIP-SAM}, achieves state-of-the-art results on our fashion domain dataset as well as standard benchmarks repurposed for our application.
  \item We propose \textbf{CtrlVTON}, a framework that enables pixel-level control over garment style, size, and placement via segmentation masks. CtrlVTON handles diverse garment categories and garment display formats, and unifies standard garment swap, layering, selective switching, and multi-garment try-on within one framework.
  \item We develop a data pipeline that synthesizes (person, garment, person-with-different-garment) triplets \emph{along with their corresponding masks}. We publicly release \textbf{VITON-HD-edit} (Sec.~\ref{sec:cvton_data}), a benchmark built from this pipeline that supports image-editing VTO, mask-controllable VTO, and instance-level visual-prompt segmentation.
\end{enumerate}

\section{Related Work}
\label{sec:related}

\subsection{Virtual Try-On}

Image-based virtual try-on (VTO) aims to synthesize a photorealistic image of a person wearing a given query garment.
Early methods geometrically warped the garment features and used GAN for rendering~\cite{han2018viton,wang2018cpvton,yang2020acgpn,lee2022hrviton,xie2023gpvton}.
However, these methods struggled with complex poses, fine textures, and occlusion.
Diffusion-based VTO substantially improved realism and detail fidelity.
Initially, diffusion-based VTO utilized latent diffusion and parallel UNets~\cite{zhu2023tryondiffusion,morelli2023ladivton}. 
Subsequent methods refined the formulation with semantic-correspondence mechanisms, improved conditioning, and architectural simplification~\cite{kim2024stableviton,choi2024idmvton,chong2024catvton,zhou2024leffa}.
More recent approaches adopt Diffusion Transformer (DiT) backbones, exploiting stronger generative priors offered by large-scale pretrained models~\cite{jiang2024fitdit,lee2025voost,chong2025fastfit,feng2025omnitry,guo2025any2anytryon}.

Many of these methods are based on image inpainting~\cite{kim2024stableviton,choi2024idmvton,chong2024catvton,jiang2024fitdit,zhou2024leffa,lee2025voost,deria2025mugavton,chong2025fastfit}.
Although inpainting enables the user to specify which region to edit, the inpainting mask is also the origin of the well-known difficulties with complex poses, occlusions, and identity drift.

These difficulties are resolved by recent editing-based methods~\cite{feng2025omnitry,guo2025any2anytryon}, which remove the reliance on inpainting masks.
Another line of work extends VTO from supporting a single reference garment to multiple garments~\cite{zhu2024mmvto,deria2025mugavton,chong2025fastfit}.
At the same time, large proprietary image editing models~\cite{google2025nanobananapro,openai2025gptimage,bytedance2025seedream4,labs2025flux1kontextflowmatching,black2025flux2}, though not designed for VTO, already match or surpass methods designed for VTO.

Although these image editing models removed the limitations of the inpainting models, they also lost the spatial control.
The user is left with no way to specify \emph{how} a garment is worn.
M\&M~VTO~\cite{zhu2024mmvto} and PromptDresser~\cite{kim2025promptdresser} come close by offering coarse text-guided layout for multi-garment outfits, but cannot express precise spatial placement and do not support layered garments.
CtrlVTON closes this gap by accepting segmentation masks within the editing framework, recovering the inpainting model's pixel-level spatial control while avoiding its other limitations.

\subsection{Controllable Image Editing}

Existing controllable image editing methods provide spatial control through various interfaces.
Structural-cue conditioning~\cite{zhang2023controlnet,shi2024dragdiffusion} generates the whole image from edges, depth, or drag points.
Layout-based methods~\cite{li2023gligen,wang2024instancediffusion,wang2024msdiffusion} operate on bounding boxes or other coarse primitives, with multi-subject variants such as MS-Diffusion~\cite{wang2024msdiffusion} pairing each box with a separate reference image.
Mask-based inpainting~\cite{rombach2022ldm,xie2023smartbrush,zhuang2024powerpaint,ju2024brushnet} provides precise spatial control through masks, with text prompts specifying the content within.
Reference-guided composition~\cite{yang2023paintbyexample,chen2024anydoor} places a reference image at a user-specified location, reproducing its appearance with high fidelity.
None of these interfaces, however, simultaneously supports reference-image fidelity, user-specified spatial layout, and identity preservation of the input person—the combination VTO requires.

\begin{figure}[t]
\centering
\includegraphics[width=\linewidth]{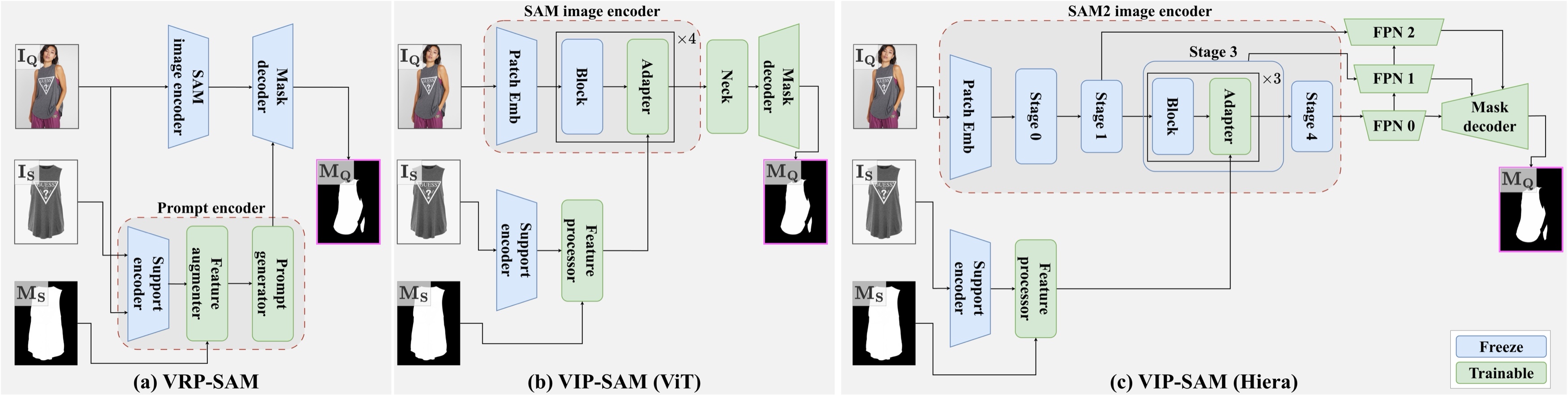}
\caption{\textbf{Segmentation model architectures.} (a) VRP-SAM trains extra modules to extract a visual prompt from support and query images; this prompt is fed to SAM's mask decoder in place of the usual spatial prompts (point, box, mask) (b, c). VIP-SAM with the SAM (ViT) encoder in (b) and the SAM2 (Hiera) encoder in (c). In each variant, adapters inject support-image features into the encoder after every block.
The output is decoded into a mask by a module trained from scratch.}
\label{fig:seg_method}
\end{figure}

\subsection{Visual-Reference-Prompt Segmentation}

The Segment Anything Model (SAM)~\cite{kirillov2023sam} is a foundation model for interactive segmentation, producing class-agnostic masks from spatial prompts (points, boxes, or coarse masks).
SAM2~\cite{ravi2024sam2} extends this paradigm to video, propagating object identities across frames through a memory module.
Both are restricted to spatial prompts and offer no mechanism for specifying a target instance through a separate reference image.
Follow-up work fills this gap with visual reference prompting: given a query and an annotated support image, the model segments the referenced object in the query.
Training-free methods like PerSAM~\cite{zhang2023personalize} and Matcher~\cite{liu2023matcher} prompt SAM with feature similarities derived from pretrained foundation models.
Training-based methods like VRP-SAM~\cite{sun2024vrpsam} and ProSAM~\cite{wang2025prosam} train a dedicated encoder that converts the reference into a SAM-compatible prompt.
More recently, SAM3~\cite{carion2025sam3} extends SAM with text and image-exemplar prompts (positive or negative bounding boxes) for concept-level segmentation.

Architectural differences aside, these methods (with the exception of SAM3) rely on feature-level correspondence between reference and query, and are evaluated at the category-level on splits like Pascal-$5^i$~\cite{shaban2017oneshot} and COCO-$20^i$~\cite{nguyen2019fwb}, or open-vocabulary benchmarks like LVIS~\cite{gupta2019lvis} and SA-Co~\cite{carion2025sam3}.
This conflates instances of the same category, whereas our setting demands instance-level discrimination: singling out a specific garment among same-category items (e.g. layered shirts of similar color).
VIP-SAM is designed to solve this problem.
\section{Visual-Instance-Prompt Segmentation (VIP-Seg)}
\label{sec:vipseg}

\noindent\textbf{Task definition.}
We define \emph{VIP-Seg} as follows: given a support image ($I_S$) of an object (\eg, a studio flatlay of a particular garment), segment precisely \emph{that instance} in a query image ($I_Q$) containing it (\eg, a person wearing that garment).
This task is distinct from VRP-Seg: the objective is to segment the exact same instance, not just any object of the same class, even in the presence of same-class distractors (\eg, a shirt worn over another shirt of similar color).
Rather than generalizing to novel classes, we focus on identifying \emph{the same instance} across scenes, under same-category distractors, heavy occlusion, and non-rigid deformation between the studio flatlay and the on-person image, typical of product detail page (PDP) images in e-commerce.

\noindent\textbf{Why prior architectures struggle.}
To see why instance-level discrimination is hard for prior methods, consider VRP-SAM's architecture in Fig.~\ref{fig:seg_method}(a).
The query image is encoded by a frozen SAM backbone, while the support image is processed in parallel to produce a visual reference prompt for the mask decoder.
For this to succeed, two conditions must hold: the frozen image encoder must disambiguate all relevant objects, and the matching modules (prompt encoder and mask decoder) must establish correspondence between objects in the support and query images.
Both conditions are difficult to meet when garments are layered or share similar color.
Empirically, VRP-SAM and similar methods struggle on exactly these cases, see Fig.~\ref{fig:seg_comparison}.

\noindent\textbf{VIP-SAM.}
We inject reference features into the backbone at earlier stages, so that the query encoder is conditioned on the support image from the outset rather than only at the prompt level.
The query image is processed by the SAM backbone (ViT) or the SAM2 backbone (Hiera), giving rise to two variants shown in Fig.~\ref{fig:seg_method}\,(b) and (c). 
The support image is processed by interchangeable encoders such as ResNet-50~\cite{he2016resnet}, DINOv2~\cite{oquab2024dinov2}, or DINOv3~\cite{simeoni2025dinov3}.
Importantly, the query image features are conditioned on the support image features by cross-attention adapters at intermediate stages of the SAM or SAM2 encoders.

\noindent\textbf{Training data.}
We construct a fashion segmentation dataset for VIP-Seg by collecting (garment, person) image pairs across diverse garment categories.
For each pair we annotate the region the garment occupies on the person image, yielding a (garment, person, mask) triplet.
The full dataset comprises roughly 10K human-labeled masks.

\noindent\textbf{Bridge to CtrlVTON.}
VIP-SAM is the prerequisite that unlocks the controllable VTO framework in Sec.~\ref{sec:cvton}.
Training CtrlVTON to learn spatial control requires accurate, garment-instance-level masks, yet existing segmentation methods cannot reliably isolate the garment on the person that matches a given visual reference (Fig.~\ref{fig:seg_comparison}).
VIP-SAM fills this gap, providing the precise masks needed to train CtrlVTON.

\section{Controllable Virtual Try-On}
\label{sec:cvton}

CtrlVTON combines the strengths of two VTO paradigms surveyed in Sec.~\ref{sec:related}: mask-conditioned \emph{inpainting} and mask-free \emph{editing}.
From editing models we adopt \emph{full-image conditioning}: rather than erasing and refilling a region, CtrlVTON is conditioned on the entire input image.
Under the right training scheme, this enables \emph{selective} information transfer: pose, identity, other garments, accessories, and background flow through from the input, while only the targeted region is modified to match the new garment.
From inpainting models we adopt \emph{mask conditioning}, which provides \emph{local} spatial control by specifying exactly where the new garment should appear.

\subsection{From Inpainting to Editing}
\label{sec:cvton_formulation}

VTO aims to produce a target person image $p$ depicting the input person wearing the garment shown in a reference image $g_{\text{ref}}$.
Inpainting-based VTO methods cast this as a masked-image completion problem $(p_{\text{masked}}, g_{\text{ref}}) \to p$, where $p_{\text{masked}}$ is obtained by erasing the garment region of $p$.

\noindent\textbf{Editing formulation.}
As discussed in Sec.~\ref{sec:related} (see also Sec.~\ref{sec:supp_inpainting_failures} of Supp.\ for examples),
the inpainting formulation requires a fragile trade-off between mask tightness and identity preservation: an undersized mask leaks residual garment pixels that corrupt the output, while an oversized mask erases identity cues the model must then hallucinate.
Even worse, the output is further biased by the shape of the mask and the contextual bias outside the mask (\eg, shadows).
We avoid this trade-off by casting VTO as an \emph{image editing} problem $(p_{\text{ref}}, g_{\text{ref}}) \to p$, where $p_{\text{ref}}$ is a reference person image depicting the same person, pose, and background as $p$ but wearing a \emph{different} garment than $g_{\text{ref}}$.
Under this formulation the model is free to modify the entire image but learns to preserve identity, pose, and background from the reference image $p_{\text{ref}}$ because of how data is prepared.
Training therefore requires triplets $(p_{\text{ref}}, g_{\text{ref}}, p)$, which no public dataset provides.
We describe how such a dataset can be constructed in Sec.~\ref{sec:cvton_data}.

\subsection{Data Curation}
\label{sec:cvton_data}

\begin{figure}[t]
\centering
\includegraphics[width=\linewidth]{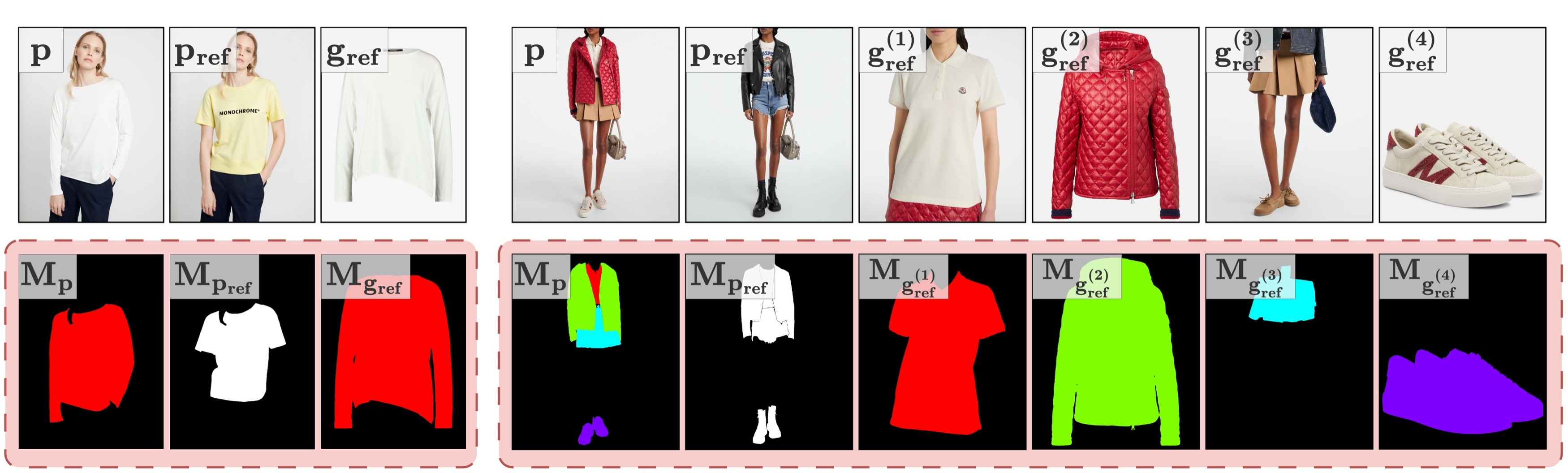}
\caption{\textbf{Overview of training samples.} We illustrate both single-garment and multi-garment data as well as the notation used throughout this work.
\textbf{Top:} target person $p$, synthetic reference person $p_{\text{ref}}$, and the reference garment(s) $g_{\text{ref}}$ (or $g_{\text{ref}}^{(1..K)}$ in the multi-garment case).
\textbf{Bottom:} the corresponding masks $M_p$, $M_{p_{\text{ref}}}$, and $M_{g_{\text{ref}}}$ provide spatial control.
In the multi-garment setting, each garment mask $M_{g_{\text{ref}}}^{(k)}$ and its corresponding region in $M_p$ are rendered in the same color, enabling simultaneous per-garment control.
}
\label{fig:data_samples}
\end{figure}

Training the editing model requires triplets $(p, p_{\text{ref}}, g_{\text{ref}})$. Source pairs $(p, g_{\text{ref}})$ are readily available from existing data sources, but $p_{\text{ref}}$ is not, so we synthesize $p_{\text{ref}}$ as described below.
For controllability, we use the three masks ($M_p$, $M_{p_{\text{ref}}}$, $M_{g_{\text{ref}}}$) corresponding to the garment regions in the three images.
Thus, each training instance is the tuple $\bigl(p,\; p_{\text{ref}},\; g_{\text{ref}},\; M_p,\; M_{p_{\text{ref}}},\; M_{g_{\text{ref}}}\bigr)$.
Fig.~\ref{fig:data_samples} shows sample training instances for both single- and multi-garment scenarios.
The remainder of this section describes how these images and masks are constructed.

\noindent\textbf{Source pairs $(p, g_{\text{ref}})$.}
We assemble person--garment pairs from three complementary sources---public VTO datasets, commercial datasets licensed from fashion retailers, and in-house datasets---spanning diverse garment categories (tops, bottoms, full-body garments, shoes, bags) and garment image formats (flatlay, on-person, in-the-wild).
A detailed breakdown of the training corpus by garment cardinality (single- vs.\ multi-garment samples), garment category, and task-token assignment is given in Sec.~\ref{sec:supp_data_sources} of Supp.

\noindent\textbf{Masks $M_p$ and $M_{g_{\text{ref}}}$.}
$M_p$ is the mask of $g_{\text{ref}}$ in $p$, obtained by querying VIP-SAM (Sec.~\ref{sec:vipseg}) with $g_{\text{ref}}$ as the visual prompt.
$M_{g_{\text{ref}}}$ is the mask of the reference garment in $g_{\text{ref}}$, obtained via dichotomous image segmentation~\cite{meyer2025ben} for flatlay images and via VIP-SAM for human-worn images.

\noindent\textbf{Synthesizing $p_{\text{ref}}$.}
Inspired by the synthetic-data construction pipelines used to train recent image-editing models~\cite{sheynin2024emuedit,wu2025qwenimage}, we synthesize $p_{\text{ref}}$ with off-the-shelf image generation models~\cite{flux2024,lee2025voost,labs2025flux1kontextflowmatching,black2025flux2}.
Given $(p, M_p, g_{\text{unpaired}})$, where $g_{\text{unpaired}}$ is a different garment that the person could plausibly wear, an inpainting system is used to produce $p_{\text{ref}}$ by filling the masked region of $p$ with $g_{\text{unpaired}}$, while preserving pose, identity, and background.
See Sec.~\ref{sec:supp_masking} of Supp.\ for full details.

\noindent\textbf{Extracting $M_{p_{\text{ref}}}$.}
$M_{p_{\text{ref}}}$ is the mask of the garment in $p_{\text{ref}}$ that must be replaced by $g_{\text{ref}}$.
For training data, it is obtained by querying VIP-SAM with $g_{\text{unpaired}}$ as the visual prompt.

\noindent\textbf{Quality control.}
For each source pair, we synthesize four to five candidates and select one via a three-stage process: VLM-based screening, contour-based filtering using the VIP-SAM masks $M_{p_{\text{ref}}}$ and $M_p$ to detect silhouette leakage, and final review by three annotators (Sec.~\ref{sec:supp_qc} of Supp.).

\noindent\textbf{VITON-HD-edit.}
Applying the data preparation pipeline to the full VITON-HD test set ($2{,}032$ images) yields VITON-HD-edit, a public benchmark supporting image-editing VTO, VIP-Seg, and spatially-controllable VTO.

\subsection{Model}
\label{sec:cvton_model}

\begin{figure}[t]
\centering
\includegraphics[width=\linewidth]{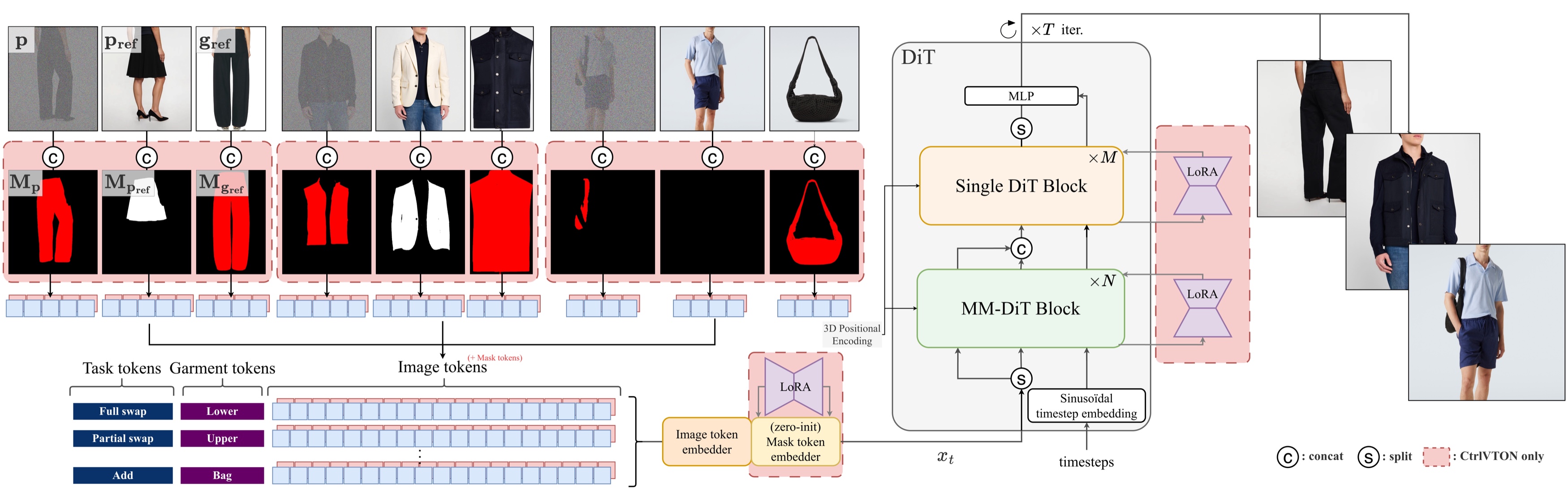}
\caption{\textbf{Overview of CtrlVTON.}
For the base model, we fine-tune an image-editing DiT backbone on triplets $(p, p_{\text{ref}}, g_{\text{ref}})$.
The controllability extension (red-dashed region) takes as input the masks $M_p$, $M_{p_{\text{ref}}}$, and $M_{g_{\text{ref}}}$, which are channel-wise concatenated with the tokens of the corresponding images.
It is implemented as a LoRA adapter on top of the frozen base model.}
\label{fig:method}
\end{figure}

We train two editing models that differ in the granularity of control.
\textbf{CtrlVTON-base} is trained on triplets $(p,\; p_{\text{ref}},\; g_{\text{ref}})$ under \emph{semantic-level} control.
Given $p_{\text{ref}}$ and $g_{\text{ref}}$, together with a garment-class token and a task token that specify the garment type and try-on operation, the model generates the final image $p$.
\textbf{CtrlVTON} extends CtrlVTON-base with \emph{pixel-level} spatial control by training a lightweight LoRA adapter on top of CtrlVTON-base, which additionally consumes the three masks $M_p$, $M_{p_{\text{ref}}}$, and $M_{g_{\text{ref}}}$.
The full architecture of both models is summarized in Fig.~\ref{fig:method}; we now describe each in turn.

\noindent\textbf{{CtrlVTON-base.}}
We fine-tune a pre-trained image-editing diffusion transformer~\cite{labs2025flux1kontextflowmatching} on triplets $(p_{\text{ref}}, g_{\text{ref}}, p)$.
To express the full range of try-on operations, the model is conditioned on two discrete semantic tokens.
The \emph{garment-class token} ($\tau_{cls}$) $\in \{\textsc{upper},\allowbreak \textsc{lower},\allowbreak \textsc{full},\allowbreak \textsc{shoes},\allowbreak \textsc{bag}\}$ specifies the type of $g_{\text{ref}}$.
The \emph{task token} ($\tau_{task}$) $\in \{\textsc{full\_swap},\allowbreak \textsc{partial\_swap},\allowbreak \textsc{add}\}$ specifies how existing garments in $p_{\text{ref}}$ should be handled:
\textsc{full\_swap} replaces all garments of the matching class with $g_{\text{ref}}$;
\textsc{partial\_swap} replaces only a single garment of that class (\eg, switching just the inner shirt while keeping the outer jacket);
and \textsc{add} preserves existing garments and places $g_{\text{ref}}$ on top (\eg, layering a jacket over a shirt).
Together, the garment-class and task tokens enable the base model to support the garment swapping, selective switching, and layering scenarios introduced in Sec.~\ref{sec:intro}.

\noindent\textbf{{CtrlVTON}}
CtrlVTON-base is an editing model that provides semantic-level control over VTO: the two discrete tokens specify \emph{which} try-on operation to perform, while the model itself determines \emph{where} the new garment should appear on the body.
CtrlVTON extends the base model with pixel-level spatial control by conditioning on the three masks $M_p$, $M_{p_{\text{ref}}}$, and $M_{g_{\text{ref}}}$ as additional inputs to the network.
It is implemented by training a LoRA adapter on top of the frozen base model, with the masks provided as additional inputs (Fig.~\ref{fig:method}).

\noindent\textbf{Inference-time mask.}
During training, $M_p$ is extracted by VIP-SAM from $(p, g_{\text{ref}})$.
During inference, $M_p$ is provided by the user. A typical workflow first runs CtrlVTON-base to generate an initial try-on result, extracts the corresponding garment mask using VIP-SAM, and then edits the mask to adjust garment style, fit, or placement before running CtrlVTON. This procedure keeps the inference-time masks aligned with the training distribution.
$M_{g_{\text{ref}}}$ is extracted by BEN2~\cite{meyer2025ben} or VIP-SAM as in training.
$M_{p_{\text{ref}}}$ can be an all-white mask (no spatial constraint), an all-black mask (no replacement, as in \textsc{add}), or a SAM-derived or hand-crafted mask when finer control is desired.

\noindent\textbf{Mask injection.}
Each mask is spatially aligned with the corresponding image by construction, so we inject it via channel-wise concatenation in the latent space of the DiT.
{Let $z_X \in \mathbb{R}^{H \times W \times C}$ and $z_{M_X} \in \mathbb{R}^{H \times W \times C}$ denote the VAE-encoded latents of image $X$ and its mask $M_X$, respectively.
For each of the three image inputs we form an augmented latent
\begin{equation}
\label{eq:mask_inject}
\tilde{z}_{X} = \bigl[\, z_{X} \;\Vert\; z_{M_X} \,\bigr], \qquad X \in \{p,\; p_{\text{ref}},\; g_{\text{ref}}\},
\end{equation}
where $\Vert$ denotes channel-wise concatenation.
This doubles the channel dimension of the latents while leaving the $H \times W$ token grid untouched, so neither the attention cost nor the conditioning-token count grows; we compare against the token-wise injection alternative~\cite{kim2026coral} in Sec.~\ref{sec:supp_injection} of Supp.

\noindent\textbf{Training objective.}
The base model is a flow-matching DiT that parameterizes the velocity field $v_\theta(x_t, t \mid c)$~\cite{labs2025flux1kontextflowmatching}.
We freeze $\theta$ and train LoRA parameters $\Delta\theta$ attached to the linear projection blocks of both the MM-DiT and Single-DiT blocks.
We optimize the standard flow-matching objective with augmented conditioning $\tilde{c} = (z_{M_{p}}, \tilde{z}_{p_{\text{ref}}}, \tilde{z}_{g_{\text{ref}}}, \tau_{\text{cls}}, \tau_{\text{task}})$:
\begin{equation}
\label{eq:ctrl_loss}
\mathcal{L}_{\text{ctrl}} = \mathbb{E}_{\substack{x_1 \sim p_{\text{data}} \\ x_0 \sim \mathcal{N}(0, I) \\ t \sim \mathcal{U}[0, 1]}}\!\left[\,\bigl\| v_{\theta + \Delta\theta}\ \!\bigl(x_t, t \mid \tilde{c}\bigr) - (x_1 - x_0) \bigr\|_2^2\,\right],
\end{equation}
where $x_t = (1-t)\,x_0 + t\,x_1$ and $x_1$ is the latent of the target $p$.

\noindent\textbf{Extension to multiple garments.}
In practical applications, outfits often consist of multiple garments, so we extend the model to jointly condition on multiple reference garments.
No architectural change is required: the model now ingests multiple reference garments $\{g_{\text{ref}}^{(1)}, \dots, g_{\text{ref}}^{(K)}\}$ instead of just one.
To preserve per-garment controllability, we color-code the conditioning masks: each garment is assigned a distinct RGB color, shared between $M_p$ and its corresponding $M_{g_{\text{ref}}^{(k)}}$.
These RGB masks are channel-wise concatenated just as in the single-garment case.
The model is trained in a single stage on the combined single- and multi-garment dataset.
All experiments in Sec.~\ref{sec:eval} use the multi-garment model.
\section{Experiments}
\label{sec:eval}

\subsection{VIP-SAM}
\label{sec:exp_vipsam}

\begin{figure}[h]
\centering
\includegraphics[width=\linewidth]{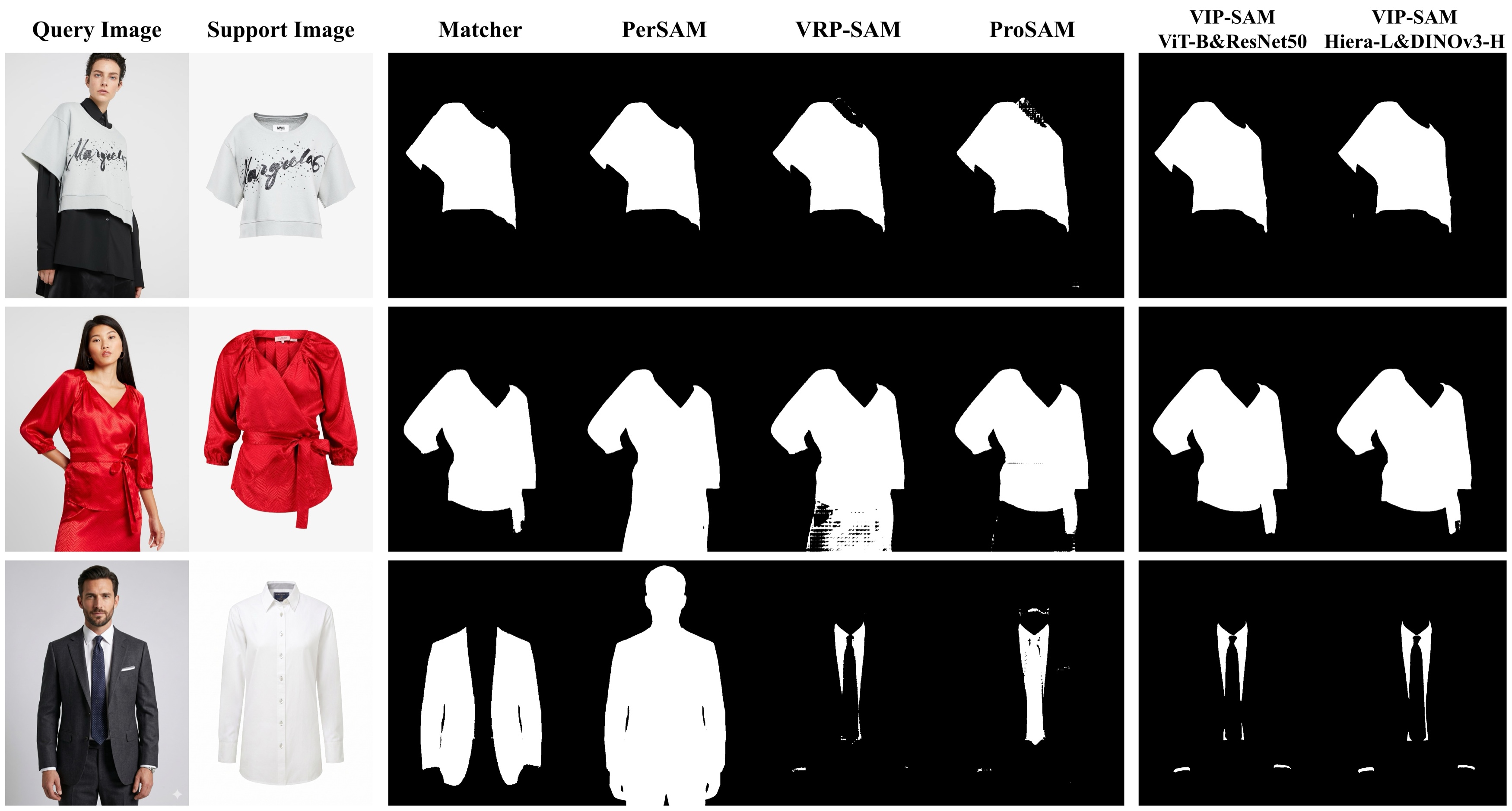}
\caption{\textbf{Qualitative comparison of visual-reference segmentation methods.} Each column shows a different method applied to the same inputs. Existing methods often include surrounding garments or the entire body, whereas VIP-SAM consistently isolates only the queried instance.}
\label{fig:seg_comparison}
\end{figure}

\begin{table}[b]
\centering
\caption{VIP-Seg results: mIoU (FB-IoU). Note that VIP-SAM (ViT-B/ResNet-50) significantly outperforms VRP-SAM (ViT-B/ResNet-50).
}
\label{tab:comparison_miou}
\setlength{\tabcolsep}{4pt}
\resizebox{\linewidth}{!}{%
\begin{tabular}{lllllll}
\toprule
Method   & Query Enc & Support Enc & Fashion-val & Fashion-test & COCO-$20^i$ & PASCAL-$5^i$ \\
\midrule
PerSAM~\cite{zhang2023personalize}   & ViT-H    & ViT-H     & 52.3 (70.2) & 47.5 (67.3) & 22.6 (56.5) & 45.5 (66.9) \\
Matcher~\cite{liu2023matcher}        & ViT-H    & DINOv2-L  & 56.8 (73.2) & 52.9 (70.8) & 50.7 (72.9) & 64.1 (78.4) \\
VRP-SAM~\cite{sun2024vrpsam}         & ViT-B    & ResNet-50 & 91.3 (95.0) & 91.4 (95.1) & 48.5 (71.9) & 56.8 (71.6) \\
VRP-SAM~\cite{sun2024vrpsam}         & ViT-H    & ResNet-50 & 94.4 (96.8) & 94.3 (96.8) & 59.4 (78.0) & 68.7 (80.5) \\
ProSAM~\cite{wang2025prosam}         & ViT-H    & ResNet-50 & 94.6 (96.9) & 93.8 (96.5) & 56.8 (76.6) & 69.3 (80.8) \\
\midrule
\rowcolor{ourRow}
VIP-SAM  & ViT-B    & ResNet-50 & 95.5 (97.4) & 95.3 (97.3) & 62.3 (79.5) & 69.4 (80.8) \\
\rowcolor{ourRow}
VIP-SAM  & Hiera-B+ & ResNet-50 & 95.8 (97.6) & 95.4 (97.4) & 63.7 (80.1) & 72.5 (83.1) \\
\rowcolor{ourRow}
VIP-SAM  & Hiera-L  & ResNet-50 & 96.5 (98.0) & 95.8 (97.6) & 67.2 (82.6) & 76.6 (85.9) \\
\rowcolor{ourRow}
VIP-SAM  & Hiera-L  & DINOv3-H  & \textbf{97.2 (98.4)} & \textbf{96.6 (98.1)} & \textbf{74.0 (86.4)} & \textbf{79.1 (87.6)} \\
\bottomrule
\end{tabular}%
}
\end{table}

We evaluate VIP-SAM on our fashion segmentation benchmark and on the standard category-level benchmarks COCO-$20^i$~\cite{nguyen2019fwb} and PASCAL-$5^i$~\cite{shaban2017oneshot}.
Since our goal is to evaluate VIP-Seg, we modify the evaluation protocol on COCO-$20^i$ and PASCAL-$5^i$ by training and evaluating on the same category, rather than on held-out classes.
Across all three evaluations, VIP-SAM achieves state-of-the-art performance (Tab.~\ref{tab:comparison_miou}), confirming that early-stage feature injection benefits instance-level segmentation.
In Fig.~\ref{fig:seg_comparison}, we show examples of layered and similar-textured garments where VRP-SAM-style late feature matching methods fail, whereas VIP-SAM correctly segments the queried instance.
Further details on training, evaluation, and compute resources are in Sec.~\ref{sec:supp_vipsam_resources} of Supp.

\subsection{CtrlVTON Experimental Setup}
\label{sec:exp_setup}

\noindent\textbf{Datasets.}
We evaluate on four public benchmarks, each chosen to probe a distinct capability of CtrlVTON.
For single-garment VTO we use \emph{VITON-HD}~\cite{choi2021vitonhd} together with a clothes-only subset of \emph{OmniTry Bench}~\cite{feng2025omnitry} containing 2{,}250 samples.
The clothes-only restriction enables fair comparison with prior single-garment VTO methods.
For multi-garment try-on, where the model must dress a person using several reference garments, we use \emph{DressCode-MR}~\cite{chong2025fastfit} and \emph{Garments2Look}~\cite{hu2026garments2look}.
For single-garment mask control, we use \emph{VITON-HD-edit} (introduced in Sec.~\ref{sec:cvton_data}).
Because no public benchmark is currently available for multi-garment mask control, we report only \emph{qualitative} results in that setting.

\noindent\textbf{Baselines.}
For single-garment VTO we compare against open-weight inpainting-based VTO models (IDM-VTON~\cite{choi2024idmvton}, CatVTON~\cite{chong2024catvton}, Leffa~\cite{zhou2024leffa}, Voost~\cite{lee2025voost}, CORAL~\cite{kim2026coral}) and editing-based VTO models (Any2AnyTryon~\cite{guo2025any2anytryon}, OmniTry~\cite{feng2025omnitry}).
For multi-garment VTO we compare against three open-weight models: OmniTry and FastFit~\cite{chong2025fastfit} transfer reference garments onto the input person, while BootComp~\cite{choi2025bootcomp} generates a new person image from the references.
For mask-controllable try-on (Sec.~\ref{sec:exp_ctrl}), no comparable open-weight baseline is available, so we benchmark against the strongest proprietary image-editing models: Nano Banana Pro~\cite{google2025nanobananapro}, GPT Image 1.5~\cite{openai2025gptimage}, Seedream 4.5~\cite{bytedance2025seedream4}, and FLUX.2~[pro]~\cite{black2025flux2}.

\noindent\textbf{Metrics.}
We report three families of metrics.
All VTO evaluation in this paper follows the \emph{unpaired} protocol: transferring a garment that the person is not originally wearing.
This precludes the use of metrics such as SSIM~\cite{wang2004ssim} and LPIPS~\cite{zhang2018lpips}, which require ground-truth target images that are unavailable for unpaired evaluation.
Distribution-level scores such as FID~\cite{heusel2017fid} remain computable against the real-image distribution, but are known to be poorly aligned with try-on quality: they capture global statistics while overlooking instance-level errors such as distorted textures or misplaced patterns~\cite{feng2025omnitry,li2026openvton}.
We therefore rely on the following instance-level criteria (further details in Sec.~\ref{sec:supp_metrics} of Supp.).

\noindent\textit{(a) Garment-fidelity metrics.}
Following OmniTry~\cite{feng2025omnitry}, we measure the cosine similarity between the reference garment and the generated garment crop.
This is calculated across two embedding spaces: M-DINO~\cite{caron2021dino}, which captures local geometry, and M-CLIP-I~\cite{radford2021clip}, which represents global semantics.
VIP-SAM (Sec.~\ref{sec:vipseg}) provides the per-instance garment crops this protocol requires.

\noindent\textit{(b) VLM-as-judge.}
We adopt a VLM-as-judge framework using Gemini 3.0 Flash~\cite{google2025gemini3}, building on recent VTO-specific adaptations~\cite{kim2026coral,li2026openvton,chen2026tstars}.
We define three complementary metrics: GTC (Garment Transfer Consistency) measures local and global fidelity to the reference garment; PBC (Person-Background Consistency) measures the preservation of identity, pose, held belongings, and background; and PR (Physical Realism) measures whether the garment appears physically plausible on the body (drape, occlusion, lighting consistency).
These metrics refine or follow established rubrics from CORAL~\cite{kim2026coral} and T-Stars-Tryon 1.0~\cite{chen2026tstars}.

\noindent\textit{(c) Mask-adherence} metrics.
For evaluating spatial-controllability (Sec.~\ref{sec:exp_ctrl}), we extract the generated mask $M_{\text{gen}}$ from the output with VIP-SAM and compare it to the input control mask $M_p$ along three complementary axes (Fig.~\ref{fig:mask_metric}): \emph{IoU} (region overlap), the \emph{Hu moment distance} $d_{\text{Hu}}$~\cite{hu1962moments,yang2025fitcontroler} (global shape), and the \emph{symmetric Hausdorff distance} $d_H$~\cite{huttenlocher1993hausdorff,yang2025fitcontroler} (worst-case boundary).

\begin{figure}[t]
\centering
\includegraphics[width=\linewidth]{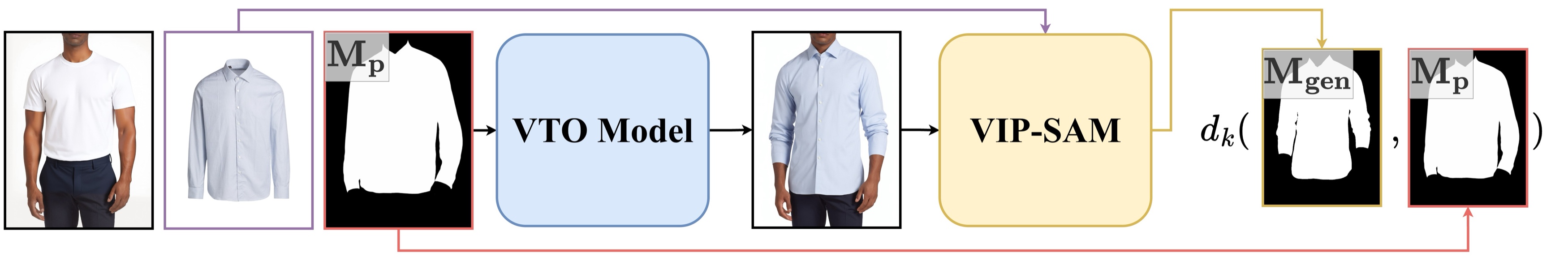}
\caption{\textbf{Illustration of the mask-adherence evaluation.} The generated mask $M_{\text{gen}}$ (re-extracted via VIP-SAM) is compared against the input control mask $M_p$ under IoU (region overlap), $d_{\text{Hu}}$ (global shape via Hu invariants), and $d_H$ (worst-case boundary deviation via symmetric Hausdorff distance).}
\label{fig:mask_metric}
\end{figure}

\subsection{Single- and Multi-Garment Virtual Try-On}
\label{sec:exp_vto}

We evaluate CtrlVTON-base on three settings: single-garment, multi-garment, and task-token control. The same checkpoint is used for all three (Sec.~\ref{sec:cvton_model}).

\noindent\textbf{Single-garment.}
Tab.~\ref{tab:single_quan} presents results on VITON-HD and OmniTry Bench.
For qualitative comparison, see Fig.~\ref{fig:vto_qual}; further examples are in Fig.~\ref{fig:supp_qual_single} of Supp.
CtrlVTON-base outperforms inpainting-based baselines and is competitive with, or surpasses, recent editing-based models on most metrics.

\begin{table}[htb]
\centering
\caption{Evaluation of single-garment VTO on VITON-HD and OmniTry Bench. Methods are grouped by formulation (Inpainting vs.\ Editing). \textbf{Best} per column is bold, \underline{second-best} is underlined; the row highlighted in light blue is our method.
}
\label{tab:single_quan}
\setlength{\tabcolsep}{3pt}
\resizebox{\linewidth}{!}{%
\begin{tabular}{llcccccccccc}
\toprule
& & \multicolumn{5}{c}{VITON-HD} & \multicolumn{5}{c}{OmniTry Bench} \\
\cmidrule(lr){3-7}\cmidrule(lr){8-12}
Formulation & Method & M-DINO$\uparrow$ & M-CLIP-I$\uparrow$ & GTC$\uparrow$ & PBC$\uparrow$ & PR$\uparrow$ & M-DINO$\uparrow$ & M-CLIP-I$\uparrow$ & GTC$\uparrow$ & PBC$\uparrow$ & PR$\uparrow$ \\
\midrule
\multirow{5}{*}{Inpainting}
& IDM-VTON~\cite{choi2024idmvton}              & 0.7198 & 0.8134 & 3.6798 & 4.0876 & \underline{4.0882} & 0.5876 & 0.7989 & 3.0124 & 3.3251 & 3.2845 \\
& CatVTON~\cite{chong2024catvton}              & 0.6821 & 0.8076 & 3.6543 & 4.0432 & 3.9644 & 0.5744 & 0.7906 & 2.9856 & 3.2942 & 3.1423 \\
& Leffa~\cite{zhou2024leffa}                   & 0.6687 & 0.7423 & 3.2987 & 3.6543 & 3.7582 & 0.5234 & 0.7345 & 2.6543 & 3.0214 & 2.9567 \\
& Voost~\cite{lee2025voost}                    & 0.7254 & 0.8089 & 3.6587 & 4.0521 & 4.0023 & 0.5798 & 0.7956 & 3.0012 & 3.3105 & 3.2144 \\
& CORAL~\cite{kim2026coral}                       & 0.7298 & 0.8054 & 3.6612 & 4.0398 & 3.8134 & 0.5821 & 0.7923 & 2.9945 & 3.2876 & 3.1021 \\
\midrule
\multirow{3}{*}{Editing}
& Any2Any Tryon~\cite{guo2025any2anytryon}     & 0.7098 & 0.7521 & 3.3234 & 3.6987 & 3.8872 & 0.5398 & 0.7398 & 3.1245 & 3.4128 & 3.3102 \\
& OmniTry~\cite{feng2025omnitry}               & \underline{0.7421} & \underline{0.8823} & \underline{4.0876} & \underline{4.5876} & 4.0187 & \underline{0.6995} & \textbf{0.8560} & \underline{3.6542} & \underline{3.8765} & \underline{3.7421} \\
\rowcolor{ourRow}
& \textbf{CtrlVTON-base}                & \textbf{0.8054} & \textbf{0.8845} & \textbf{4.2057} & \textbf{4.7301} & \textbf{4.3753} & \textbf{0.7282} & \underline{0.8551} & \textbf{4.1401} & \textbf{4.5439} & \textbf{4.3121} \\
\bottomrule
\end{tabular}%
}
\end{table}

\begin{figure}[htb]
\centering
\includegraphics[width=\linewidth]{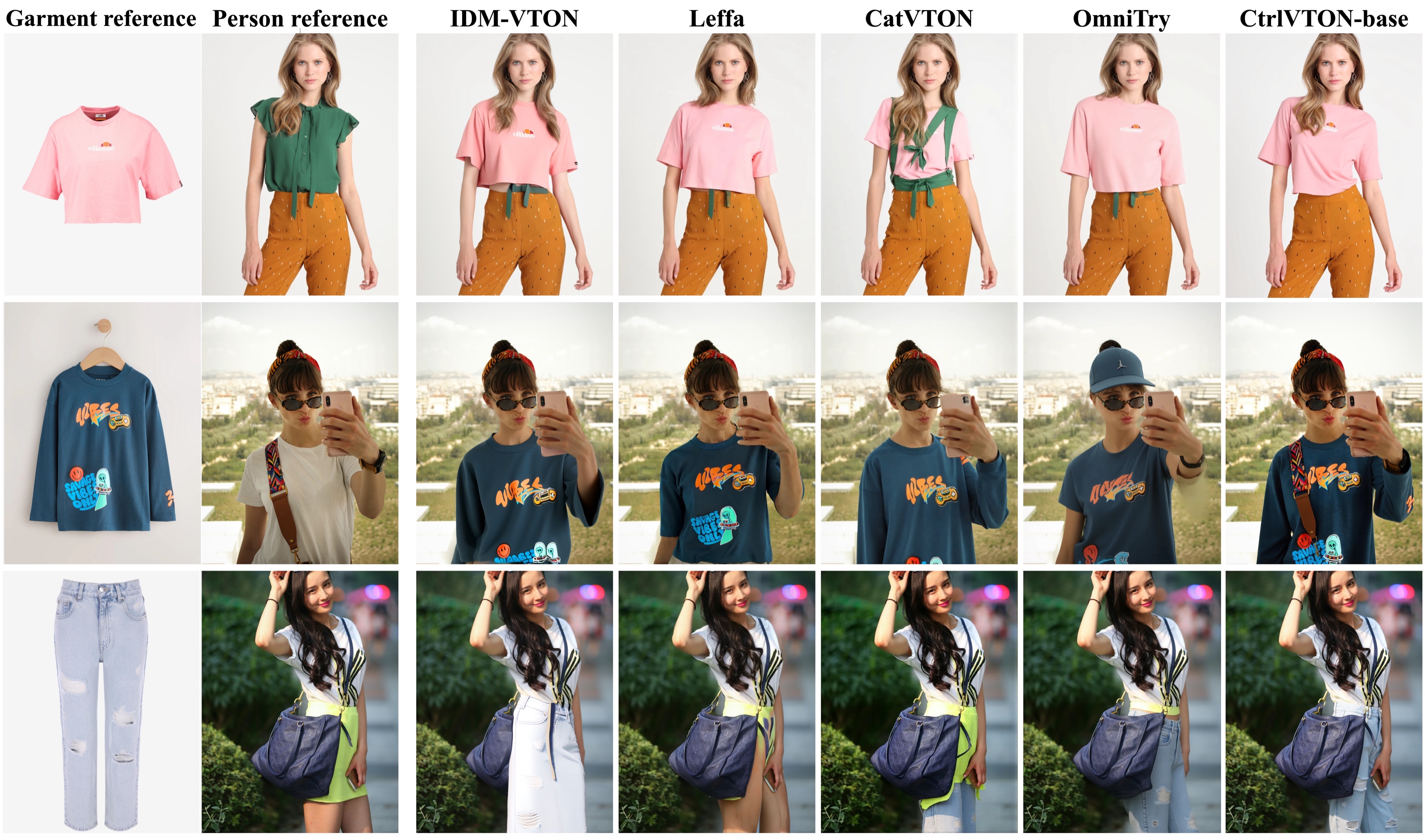}
\caption{\textbf{Qualitative comparison on single-garment VTO.}
Only \textbf{CtrlVTON-base} faithfully renders the reference garment while preserving the person's hairstyle and personal items (phone, watch, and shoulder bag).
Additional examples are provided in Fig.~\ref{fig:supp_qual_single} of Supp.}
\label{fig:vto_qual}
\end{figure}

\noindent\textbf{Multi-garment.}
Tab.~\ref{tab:multi_quan} presents results on DressCode-MR and Garments2Look (qualitative comparison in Fig.~\ref{fig:supp_qual_multi} of Supp.).
CtrlVTON-base surpasses methods using architectures designed for multi-garment (FastFit), sequential single-garment inference (OmniTry), or person regeneration from references (BootComp).

\begin{table}[ht]
\centering
\caption{Evaluation of multi-garment VTO on DressCode-MR and Garments2Look. \textbf{Best} per column is bold, \underline{second-best} is underlined; the row highlighted in light blue is our method.
``--'' denotes that PBC is not applicable to BootComp, which synthesizes a new person image from the references rather than preserving an input one. $^{\ast}$\,Garments2Look does not provide an unpaired split, so we construct one by randomly pairing persons with garments within the test set.}
\label{tab:multi_quan}
\setlength{\tabcolsep}{3pt}
\resizebox{\linewidth}{!}{%
\begin{tabular}{lcccccccccc}
\toprule
& \multicolumn{5}{c}{DressCode-MR} & \multicolumn{5}{c}{Garments2Look~$^{\ast}$} \\
\cmidrule(lr){2-6}\cmidrule(lr){7-11}
Method & M-DINO$\uparrow$ & M-CLIP-I$\uparrow$ & GTC$\uparrow$ & PBC$\uparrow$ & PR$\uparrow$ & M-DINO$\uparrow$ & M-CLIP-I$\uparrow$ & GTC$\uparrow$ & PBC$\uparrow$ & PR$\uparrow$ \\
\midrule
BootComp~\cite{choi2025bootcomp}               & 0.5412 & 0.7023 & 2.9845 & --     & 3.1245 & 0.5312 & 0.7124 & 2.8456 & --     & 2.9874 \\
OmniTry~\cite{feng2025omnitry}                & 0.5987 & 0.7456 & 3.2543 & 4.0123 & 3.4218 & 0.5543 & 0.7321 & 3.1254 & \underline{3.8942} & 3.3421 \\
FastFit~\cite{chong2025fastfit}               & \underline{0.6512} & \underline{0.8276} & \underline{3.7543} & \underline{4.1123} & \underline{3.9276} & \underline{0.6012} & \underline{0.7854} & \underline{3.4128} & 3.6543 & \underline{3.5218} \\
\midrule
\rowcolor{ourRow}
\textbf{CtrlVTON-base}                        & \textbf{0.6712} & \textbf{0.8456} & \textbf{3.8987} & \textbf{4.6856} & \textbf{4.1556} & \textbf{0.6589} & \textbf{0.8398} & \textbf{3.8743} & \textbf{4.6221} & \textbf{4.1370} \\
\bottomrule
\end{tabular}%
}
\end{table}

\noindent\textbf{Task-token control.}
The standard VTO benchmarks above target only a single task (\textsc{full\_swap}), so they do not cover the full expressivity of CtrlVTON-base.
In addition to \textsc{full\_swap}, which replaces every existing garment of the matching class with the reference, we also use \textsc{partial\_swap}, which replaces a single user-selected garment while preserving the rest, and \textsc{add}, which keeps every existing garment in place and layers the reference on top.
Fig.~\ref{fig:task_token_qual} shows the model's response to all three task tokens.
\textsc{full\_swap} replaces every existing garment of the matching class with the reference, \textsc{partial\_swap} replaces a single user-selected garment while preserving the rest, and \textsc{add} keeps every existing garment in place and layers the reference on top.
Switching the task token is the only change between these outputs, demonstrating that a single token-based interface unifies full swap, selective swap, and layering.
More qualitative and quantitative evaluations of token-following consistency are reported in Sec.~\ref{sec:supp_tfc} of Supp.

\begin{figure}[htb]
\centering
\includegraphics[width=\linewidth]{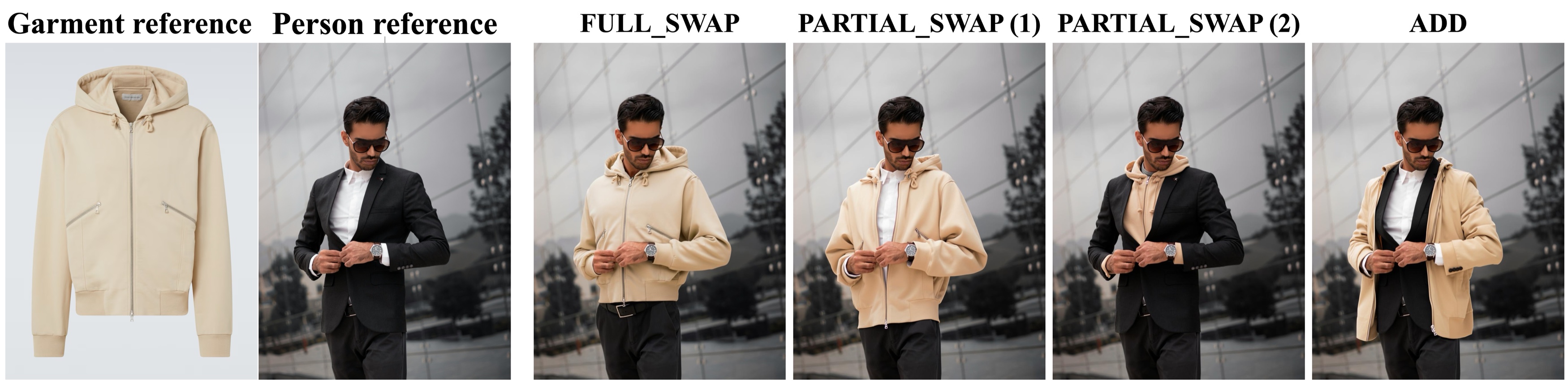}
\caption{\textbf{Effect of the task token on CtrlVTON-base results.} For the same (person, garment) pair, switching only the task token yields qualitatively different try-ons: \textsc{full\_swap} replaces every \textsc{upper}-class garment with the reference; \textsc{partial\_swap} replaces a single \textsc{upper} garment; \textsc{add} keeps all existing garments and layers the reference on top.}
\label{fig:task_token_qual}
\end{figure}

\FloatBarrier
\subsection{Mask-Controllable Try-On}
\label{sec:exp_ctrl}
To evaluate CtrlVTON's ability to faithfully place garments via user-provided masks, we benchmark our method on the VITON-HD-edit dataset against four proprietary editing models~\cite{google2025nanobananapro, openai2025gptimage, bytedance2025seedream4, black2025flux2}.
For a fair comparison, each proprietary baseline receives $p$, $g_{\text{ref}}$, and $M_p$ as three reference images, together with a text prompt that explains the role of each reference and describes the desired final image.
Tab.~\ref{tab:ctrl_quan} presents the headline result: \textbf{CtrlVTON achieves substantially better spatial control ($\mathrm{IoU}$, $d_{\text{Hu}}$, $d_H$) than the proprietary baselines by a wide margin while remaining competitive on fidelity metrics}.
While proprietary models maintain high fidelity, they exhibit a notable lack of spatial adherence, often failing to align the garment with the provided mask (Fig.~\ref{fig:metric_comparison}).

\begin{table}[ht]
\centering
\caption{Evaluation of single-garment mask-controllable VTO on VITON-HD-edit. \textbf{Best} per column is bold, \underline{second-best} is underlined; the row highlighted in light blue is our method.
}
\label{tab:ctrl_quan}
\setlength{\tabcolsep}{4pt}
\resizebox{\linewidth}{!}{%
\begin{tabular}{lcccccccc}
\toprule
Method & IoU$\uparrow$ & $d_{\text{Hu}}\downarrow$ & $d_H\downarrow$ & M-DINO$\uparrow$ & M-CLIP-I$\uparrow$ & GTC$\uparrow$ & PBC$\uparrow$ & PR$\uparrow$ \\
\midrule
Nano Banana Pro~\cite{google2025nanobananapro}    & 0.871 & 0.0044 & \underline{35.46} & \textbf{0.8256} & \textbf{0.9087} & \textbf{4.2856} & 4.1287 & \underline{4.4877} \\
GPT Image 1.5~\cite{openai2025gptimage}      & 0.811 & 0.0074 & 53.28 & 0.7854 & 0.8887 & 4.0876 & 3.5743 & 4.2910 \\
Seedream 4.5~\cite{bytedance2025seedream4}       & 0.865 & \underline{0.0039} & 41.99 & 0.8165 & \underline{0.9065} & 4.2721 & 4.6276 & 4.3788 \\
FLUX.2~[pro]~\cite{black2025flux2}         & \underline{0.873} & 0.0053 & 38.20 & 0.7943 & 0.9034 & 4.2654 & \underline{4.6589} & \textbf{4.4912} \\
\midrule
\rowcolor{ourRow}
\textbf{CtrlVTON} & \textbf{0.961} & \textbf{0.0022} & \textbf{26.05} & \underline{0.8212} & 0.9052 & \underline{4.2773} & \textbf{4.8352} & 4.4219 \\
\bottomrule
\end{tabular}%
}
\end{table}

Fig.~\ref{fig:ctrlvto_qual} surveys the controllability enabled by our system: spatial control over fit and placement, unified support for layering and selective swap, and per-garment spatial control via color-coded masks in multi-garment settings.
The mask interface gives users direct control over the degree and spatial detail of styling decisions.
This level of precision is difficult to convey via text alone: how far a shirt is tucked in, where a zipper stops, how a sleeve is rolled, or how oversized the fit should be.
The mask interface thereby establishes VTO as an interactive styling tool rather than a one-shot outfit generator.

\begin{figure}[htb]
\centering
\includegraphics[width=\linewidth]{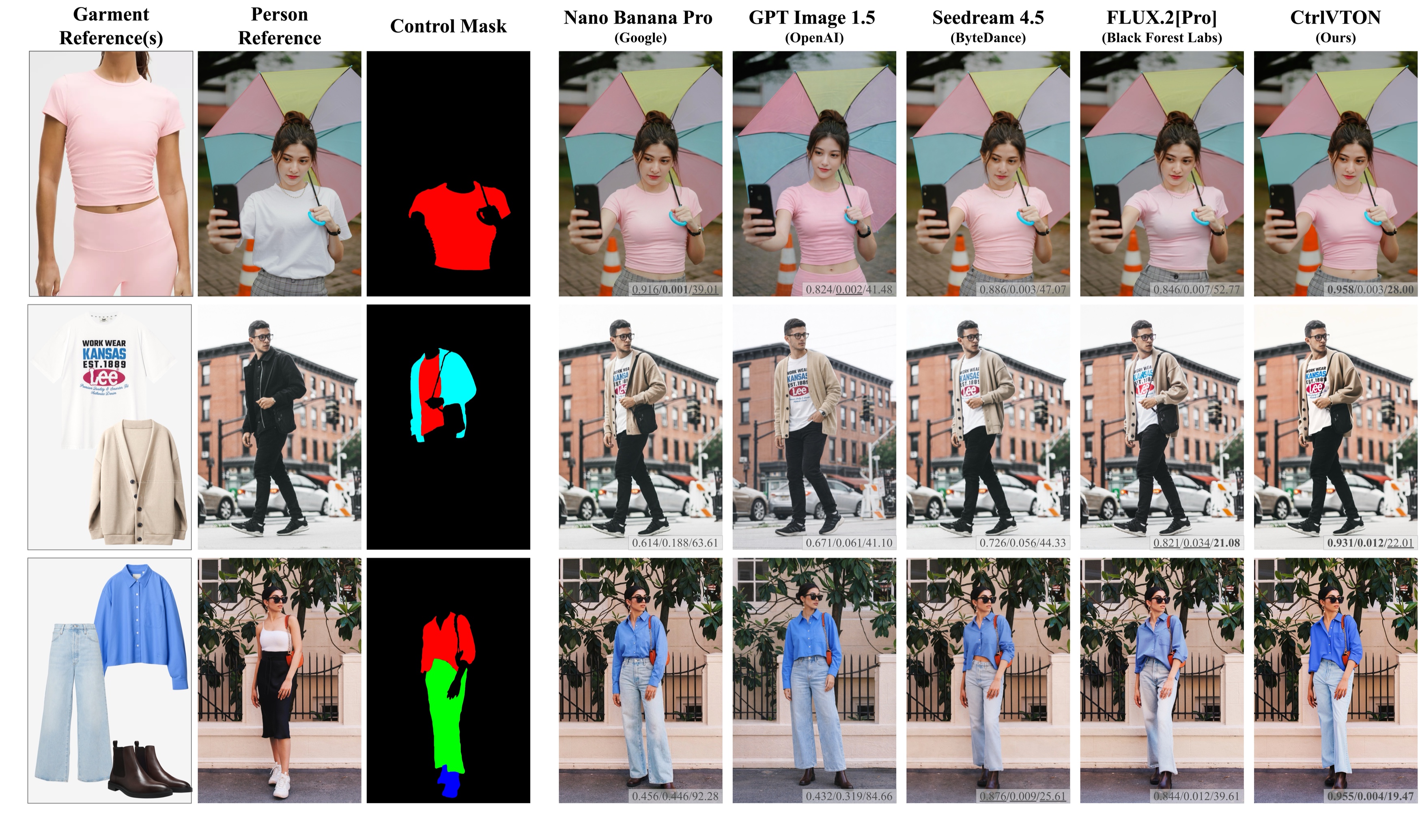}
\caption{\textbf{Qualitative comparison of mask-controllable try-on against proprietary editing models.} Both \emph{single-garment} and \emph{multi-garment} scenarios are covered. Each output is overlaid with per-image mask-adherence metrics $(IoU, d_{\text{Hu}}, d_H)$, averaged across masks for multi-garment cases. Control masks are hand-drawn by the authors.
Fig.~\ref{fig:supp_mask_metric_visual} of the appendix overlays $M_p$ on each generated image, allowing direct visual inspection of spatial agreement.}
\label{fig:metric_comparison}
\end{figure}

\begin{figure}[htb]
\centering
\includegraphics[width=\linewidth]{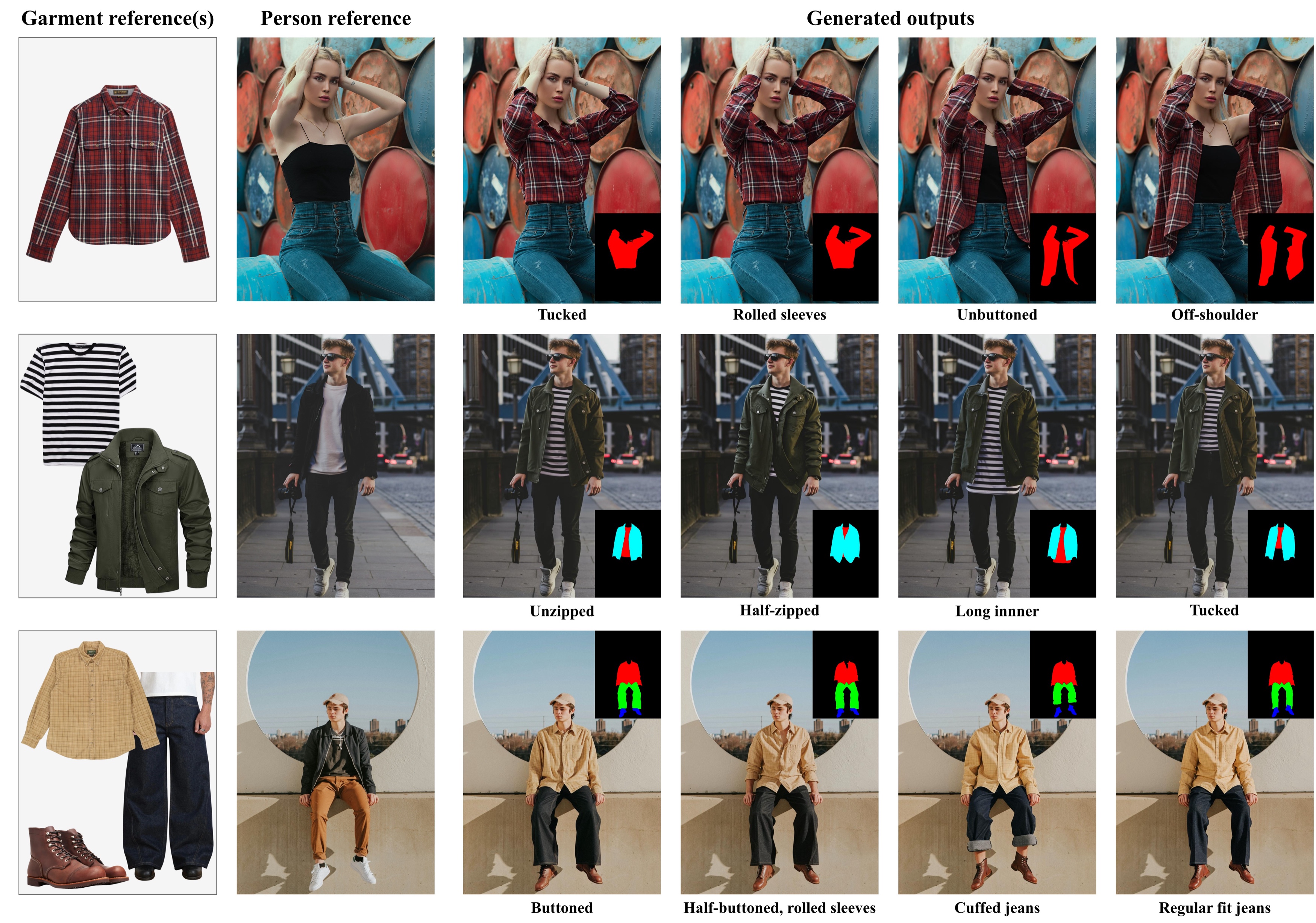}
\caption{\textbf{Fine-grained control via mask conditioning.} Given the same garment and person reference, CtrlVTON generates diverse outputs by varying only the input mask. (Top) Styling of a single garment: tucking, buttoning, and sleeve length. (Middle) Layered styling of an outer garment over an inner one: zipping, tucking, and sizing. (Bottom) Joint styling of a full outfit including top, bottom, and footwear: buttoning, cuffing, and sizing. Inset masks (right of each output) visualize the spatial conditioning and are hand-drawn by the authors. The label below each output describes the authors' intent. Additional examples are provided in Fig.~\ref{fig:supp_qual_ctrl} in the appendix.}
\label{fig:ctrlvto_qual}
\end{figure}

\section{Conclusion}
\label{sec:conclusion}

We presented \textbf{CtrlVTON}, a framework that treats controllability as a 
first-class objective in virtual try-on.
\textbf{VIP-SAM} re-identifies a specific garment across flatlay, on-person, and in-the-wild images, yielding the masks necessary for mask-conditioned VTO systems.
\textbf{CtrlVTON} recasts try-on as an editing problem over triplets $(p_{\text{ref}}, g_{\text{ref}}, p)$, sidestepping the structural limitations of the inpainting formulation while adding pixel-precise spatial control through a lightweight mask-conditioning LoRA.
The resulting model offers a unified framework for garment swapping, layering, and multi-garment composition.
To facilitate future research on editing-based and spatially-controllable VTO, we release \textbf{VITON-HD-edit} as a public testbed for both tasks.

\FloatBarrier
\clearpage

\bibliographystyle{splncs04}
\bibliography{main}

@String(PAMI  = {IEEE Trans. Pattern Anal. Mach. Intell.})

@String(CVPR  = {IEEE Conf. Comput. Vis. Pattern Recog.})

@String(ICCV  = {Int. Conf. Comput. Vis.})

@String(ECCV  = {Eur. Conf. Comput. Vis.})

@String(NeurIPS = {Adv. Neural Inform. Process. Syst.})

@String(ICML  = {Int. Conf. Mach. Learn.})

@String(ICLR  = {Int. Conf. Learn. Represent.})

@String(BMVC  = {Brit. Mach. Vis. Conf.})

@String(TMLR  = {Trans. Mach. Learn Res.})

@String(ACMMM = {ACM Int. Conf. Multimedia})

@String(PAMI  = {IEEE TPAMI})

@String(CVPR  = {CVPR})

@String(ICCV  = {ICCV})

@String(ECCV  = {ECCV})

@String(NeurIPS = {NeurIPS})

@String(ICML  = {ICML})

@String(ICLR  = {ICLR})

@String(BMVC  =	{BMVC})

@String(TMLR  = {TMLR})

@String(ACMMM = {ACM MM})

@inproceedings{han2018viton,
  author    = {Xintong Han and Zuxuan Wu and Zhe Wu and Ruichi Yu and Larry S. Davis},
  title     = {{VITON}: An Image-Based Virtual Try-On Network},
  booktitle = CVPR,
  year      = {2018}
}

@inproceedings{wang2018cpvton,
  author    = {Bochao Wang and Huabin Zheng and Xiaodan Liang and Yimin Chen and Liang Lin and Meng Yang},
  title     = {Toward Characteristic-Preserving Image-Based Virtual Try-On Network},
  booktitle = ECCV,
  year      = {2018}
}

@inproceedings{yang2020acgpn,
  author    = {Han Yang and Ruimao Zhang and Xiaobao Guo and Wei Liu and Wangmeng Zuo and Ping Luo},
  title     = {Towards Photo-Realistic Virtual Try-On by Adaptively Generating$\leftrightarrow$Preserving Image Content},
  booktitle = CVPR,
  year      = {2020}
}

@inproceedings{lee2022hrviton,
  author    = {Sangyun Lee and Gyojung Gu and Sunghyun Park and Seunghwan Choi and Jaegul Choo},
  title     = {High-Resolution Virtual Try-On with Misalignment and Occlusion-Handled Conditions},
  booktitle = ECCV,
  year      = {2022}
}

@inproceedings{xie2023gpvton,
  author    = {Zhenyu Xie and Zaiyu Huang and Xin Dong and Fuwei Zhao and Haoye Dong and Xijin Zhang and Feida Zhu and Xiaodan Liang},
  title     = {{GP-VTON}: Towards General Purpose Virtual Try-on via Collaborative Local-Flow Global-Parsing Learning},
  booktitle = CVPR,
  year      = {2023}
}

@inproceedings{zhu2023tryondiffusion,
  author    = {Luyang Zhu and Dawei Yang and Tyler Zhu and Fitsum Reda and William Chan and Chitwan Saharia and Mohammad Norouzi and Ira Kemelmacher-Shlizerman},
  title     = {{TryOnDiffusion}: A Tale of Two {UNets}},
  booktitle = CVPR,
  year      = {2023}
}

@inproceedings{morelli2023ladivton,
  author    = {Davide Morelli and Alberto Baldrati and Giuseppe Cartella and Marcella Cornia and Marco Bertini and Rita Cucchiara},
  title     = {{LaDI-VTON}: Latent Diffusion Textual-Inversion Enhanced Virtual Try-On},
  booktitle = ACMMM,
  year      = {2023}
}

@inproceedings{lee2025voost,
  author    = {Seungyong Lee and Jeong-gi Kwak},
  title     = {Voost: A Unified and Scalable Diffusion Transformer for Bidirectional Virtual Try-On and Try-Off},
  booktitle = {ACM SIGGRAPH Asia},
  year      = {2025}
}

@misc{flux2024,
  author       = {{Black Forest Labs}},
  title        = {{FLUX}},
  year         = {2024},
  howpublished = {\url{https://github.com/black-forest-labs/flux}}
}

@misc{labs2025flux1kontextflowmatching,
  author        = {{Black Forest Labs} and Stephen Batifol and Andreas Blattmann and Frederic Boesel and Saksham Consul and Cyril Diagne and Tim Dockhorn and Jack English and Zion English and Patrick Esser and Sumith Kulal and Kyle Lacey and Yam Levi and Cheng Li and Dominik Lorenz and Jonas M\"uller and Dustin Podell and Robin Rombach and Harry Saini and Axel Sauer and Luke Smith},
  title         = {{FLUX.1 Kontext}: Flow Matching for In-Context Image Generation and Editing in Latent Space},
  year          = {2025},
  eprint        = {2506.15742},
  archivePrefix = {arXiv},
  primaryClass  = {cs.GR},
  url           = {https://arxiv.org/abs/2506.15742}
}

@inproceedings{kim2024stableviton,
  author    = {Jeongho Kim and Gyojung Gu and Minho Park and Sunghyun Park and Jaegul Choo},
  title     = {{StableVITON}: Learning Semantic Correspondence with Latent Diffusion Model for Virtual Try-On},
  booktitle = CVPR,
  year      = {2024}
}

@inproceedings{choi2024idmvton,
  author    = {Yisol Choi and Sangkyung Kwak and Kyungmin Lee and Hyungwon Choi and Jinwoo Shin},
  title     = {Improving Diffusion Models for Authentic Virtual Try-on in the Wild},
  booktitle = ECCV,
  year      = {2024}
}

@inproceedings{chong2024catvton,
  author    = {Zheng Chong and Xiao Dong and Haoxiang Li and Shiyue Zhang and Wenqing Zhang and Xujie Zhang and Hanqing Zhao and Dongmei Jiang and Xiaodan Liang},
  title     = {{CatVTON}: Concatenation Is All You Need for Virtual Try-On with Diffusion Models},
  booktitle = ICLR,
  year      = {2025}
}

@misc{jiang2024fitdit,
  author        = {Boyuan Jiang and Xiaobin Hu and Donghao Luo and Qingdong He and Chengming Xu and Jinlong Peng and Jiangning Zhang and Chengjie Wang and Yunsheng Wu and Yanwei Fu},
  title         = {{FitDiT}: Advancing the Authentic Garment Details for High-fidelity Virtual Try-on},
  year          = {2024},
  eprint        = {2411.10499},
  archivePrefix = {arXiv},
  primaryClass  = {cs.CV},
  url           = {https://arxiv.org/abs/2411.10499}
}

@inproceedings{zhou2024leffa,
  author    = {Zijian Zhou and Shikun Liu and Xiao Han and Haozhe Liu and Kam Woh Ng and Tian Xie and Yuren Cong and Hang Li and Mengmeng Xu and Juan-Manuel P\'erez-Rua and Aditya Patel and Tao Xiang and Miaojing Shi and Sen He},
  title     = {Learning Flow Fields in Attention for Controllable Person Image Generation},
  booktitle = CVPR,
  year      = {2025}
}

@article{deria2025mugavton,
  author  = {Ankan Deria and others},
  title   = {{MuGa-VTON}: Multi-Garment Virtual Try-On via Diffusion Transformers with Prompt Customization},
  journal = {arXiv preprint arXiv:2508.08488},
  year    = {2025}
}

@article{bai2025qwen3,
  title={Qwen3-vl technical report},
  author={Bai, Shuai and Cai, Yuxuan and Chen, Ruizhe and Chen, Keqin and Chen, Xionghui and Cheng, Zesen and Deng, Lianghao and Ding, Wei and Gao, Chang and Ge, Chunjiang and others},
  journal={arXiv preprint arXiv:2511.21631},
  year={2025}
}

@article{chong2025fastfit,
  author  = {Zheng Chong and Yanwei Lei and Shiyue Zhang and Zhuandi He and Zhen Wang and Xujie Zhang and Xiao Dong and Yiling Wu and Dongmei Jiang and Xiaodan Liang},
  title   = {{FastFit}: Accelerating Multi-Reference Virtual Try-On via Cacheable Diffusion Models},
  journal = {arXiv preprint arXiv:2508.20586},
  year    = {2025}
}

@inproceedings{choi2025bootcomp,
  author    = {Yisol Choi and Sangkyung Kwak and Sihyun Yu and Hyungwon Choi and Jinwoo Shin},
  title     = {Controllable Human Image Generation with Personalized Multi-Garments},
  booktitle = CVPR,
  year      = {2025}
}

@inproceedings{sheynin2024emuedit,
  author    = {Shelly Sheynin and Adam Polyak and Uriel Singer and Yuval Kirstain and Amit Zohar and Oron Ashual and Devi Parikh and Yaniv Taigman},
  title     = {{Emu Edit}: Precise Image Editing via Recognition and Generation Tasks},
  booktitle = CVPR,
  year      = {2024}
}

@misc{wu2025qwenimage,
  author        = {Chenfei Wu and Jiahao Li and Jingren Zhou and Junyang Lin and Kaiyuan Gao and Kun Yan and Sheng-ming Yin and Shuai Bai and Xiao Xu and Yilei Chen and Yuxiang Chen and Zecheng Tang and Zekai Zhang and Zhengyi Wang and An Yang and Bowen Yu and Chen Cheng and Dayiheng Liu and Deqing Li and Hang Zhang and Hao Meng and Hu Wei and Jingyuan Ni and Kai Chen and Kuan Cao and Liang Peng and Lin Qu and Minggang Wu and Peng Wang and Shuting Yu and Tingkun Wen and Wensen Feng and Xiaoxiao Xu and Yi Wang and Yichang Zhang and Yongqiang Zhu and Yujia Wu and Yuxuan Cai and Zenan Liu},
  title         = {{Qwen-Image} Technical Report},
  year          = {2025},
  eprint        = {2508.02324},
  archivePrefix = {arXiv},
  primaryClass  = {cs.CV},
  url           = {https://arxiv.org/abs/2508.02324}
}

@inproceedings{kirillov2023sam,
  author    = {Alexander Kirillov and Eric Mintun and Nikhila Ravi and Hanzi Mao and Chloe Rolland and Laura Gustafson and Tete Xiao and Spencer Whitehead and Alexander C. Berg and Wan-Yen Lo and Piotr Dollar and Ross Girshick},
  title     = {Segment Anything},
  booktitle = ICCV,
  year      = {2023}
}

@inproceedings{ravi2024sam2,
  author    = {Nikhila Ravi and Valentin Gabeur and Yuan-Ting Hu and Ronghang Hu and Chaitanya Ryali and Tengyu Ma and Haitham Khedr and Roman R{\"a}dle and Chloe Rolland and Laura Gustafson and Eric Mintun and Junting Pan and Kalyan Vasudev Alwala and Nicolas Carion and Chao-Yuan Wu and Ross Girshick and Piotr Doll{\'a}r and Christoph Feichtenhofer},
  title     = {{SAM} 2: Segment Anything in Images and Videos},
  booktitle = ICLR,
  year      = {2025}
}

@article{carion2025sam3,
  author  = {Nicolas Carion and Laura Gustafson and Yuan-Ting Hu and others},
  title   = {{SAM} 3: Segment Anything with Concepts},
  journal = {arXiv preprint arXiv:2511.16719},
  year    = {2025}
}

@inproceedings{chen2024anydoor,
  author    = {Xi Chen and Lianghua Huang and Yu Liu and Yujun Shen and Deli Zhao and Hengshuang Zhao},
  title     = {{AnyDoor}: Zero-shot Object-level Image Customization},
  booktitle = CVPR,
  year      = {2024}
}

@inproceedings{yang2023paintbyexample,
  author    = {Binxin Yang and Shuai Gu and Bo Zhang and Ting Zhang and Xuejin Chen and Xiaowen Sun and Fang Wen},
  title     = {Paint by Example: Exemplar-based Image Editing with Diffusion Models},
  booktitle = CVPR,
  year      = {2023}
}

@inproceedings{zhang2023personalize,
  author    = {Renrui Zhang and Zhengkai Jiang and Ziyu Guo and Shilin Yan and Junting Pan and Hao Dong and Yu Qiao and Peng Gao and Hongsheng Li},
  title     = {Personalize Segment Anything Model with One Shot},
  booktitle = ICLR,
  year      = {2024}
}

@inproceedings{liu2023matcher,
  author    = {Yang Liu and Muzhi Zhu and Hengtao Li and Hao Chen and Xinlong Wang and Chunhua Shen},
  title     = {Matcher: Segment Anything with One Shot Using All-Purpose Feature Matching},
  booktitle = ICLR,
  year      = {2024}
}

@inproceedings{sun2024vrpsam,
  author    = {Yanpeng Sun and Jiahui Chen and Shan Zhang and Xinyu Zhang and Qiang Chen and Gang Zhang and Errui Ding and Jingdong Wang and Zechao Li},
  title     = {{VRP-SAM}: {SAM} with Visual Reference Prompt},
  booktitle = CVPR,
  year      = {2024}
}

@inproceedings{wang2025prosam,
  title={ProSAM: Enhancing the Robustness of SAM-based Visual Reference Segmentation with Probabilistic Prompts},
  author={Wang, Xiaoqi and Sebastian, Clint and He, Wenbin and Ren, Liu},
  booktitle=ICCV,
  year={2025}
}

@inproceedings{wang2023seggpt,
  author    = {Xinlong Wang and Xiaosong Zhang and Yue Cao and Wen Wang and Chunhua Shen and Tiejun Huang},
  title     = {{SegGPT}: Towards Segmenting Everything in Context},
  booktitle = ICCV,
  year      = {2023}
}

@inproceedings{wang2023images,
 title={Images speak in images: A generalist painter for in-context visual learning},
 author={Wang, Xinlong and Wang, Wen and Cao, Yue and Shen, Chunhua and Huang, Tiejun},
 booktitle=CVPR,
 year={2023}
}

@article{meyer2025ben,
  title={Ben: Using confidence-guided matting for dichotomous image segmentation},
  author={Meyer, Maxwell and Spruyt, Jack},
  journal={arXiv preprint arXiv:2501.06230},
  year={2025}
}

@inproceedings{zhu2024mmvto,
  author    = {Luyang Zhu and Yingwei Li and Nan Liu and Hao Peng and Dawei Yang and Ira Kemelmacher-Shlizerman},
  title     = {{M\&M} {VTO}: Multi-Garment Virtual Try-On and Editing},
  booktitle = CVPR,
  year      = {2024}
}

@inproceedings{kim2025promptdresser,
  title={Promptdresser: Improving the quality and controllability of virtual try-on via generative textual prompt and prompt-aware mask},
  author={Kim, Jeongho and Jin, Hoiyeong and Park, Sunghyun and Choo, Jaegul},
  booktitle={Proceedings of the IEEE/CVF International Conference on Computer Vision},
  pages={16026--16036},
  year={2025}
}

@inproceedings{zhang2023controlnet,
  author    = {Lvmin Zhang and Anyi Rao and Maneesh Agrawala},
  title     = {Adding Conditional Control to Text-to-Image Diffusion Models},
  booktitle = ICCV,
  year      = {2023}
}

@inproceedings{shi2024dragdiffusion,
  author    = {Yujun Shi and Chuhui Xue and Jun Hao Liew and Jiachun Pan and Hanshu Yan and Wenqing Zhang and Vincent Y. F. Tan and Song Bai},
  title     = {{DragDiffusion}: Harnessing Diffusion Models for Interactive Point-based Image Editing},
  booktitle = CVPR,
  year      = {2024}
}

@inproceedings{ju2024brushnet,
  author    = {Xuan Ju and Xian Liu and Xintao Wang and Yuxuan Bian and Ying Shan and Qiang Xu},
  title     = {{BrushNet}: A Plug-and-Play Image Inpainting Model with Decomposed Dual-Branch Diffusion},
  booktitle = ECCV,
  year      = {2024}
}

@inproceedings{rombach2022ldm,
  author    = {Robin Rombach and Andreas Blattmann and Dominik Lorenz and Patrick Esser and Bj{\"o}rn Ommer},
  title     = {High-Resolution Image Synthesis with Latent Diffusion Models},
  booktitle = CVPR,
  year      = {2022}
}

@inproceedings{xie2023smartbrush,
  author    = {Shaoan Xie and Zhifei Zhang and Zhe Lin and Tobias Hinz and Kun Zhang},
  title     = {{SmartBrush}: Text and Shape Guided Object Inpainting with Diffusion Model},
  booktitle = CVPR,
  year      = {2023}
}

@inproceedings{zhuang2024powerpaint,
  author    = {Junhao Zhuang and Yanhong Zeng and Wenran Liu and Chun Yuan and Kai Chen},
  title     = {A Task is Worth One Word: Learning with Task Prompts for High-Quality Versatile Image Inpainting},
  booktitle = ECCV,
  year      = {2024}
}

@inproceedings{guo2025any2anytryon,
  author    = {Hailong Guo and Bohan Zeng and Yiren Song and Wentao Zhang and Chuang Zhang and Jiaming Liu},
  title     = {{Any2AnyTryon}: Leveraging Adaptive Position Embeddings for Versatile Virtual Clothing Tasks},
  booktitle = ICCV,
  year      = {2025}
}

@inproceedings{shaban2017oneshot,
  author    = {Amirreza Shaban and Shray Bansal and Zhen Liu and Irfan Essa and Byron Boots},
  title     = {One-Shot Learning for Semantic Segmentation},
  booktitle = BMVC,
  year      = {2017},
}

@inproceedings{nguyen2019fwb,
  author    = {Khoi Nguyen and Sinisa Todorovic},
  title     = {Feature Weighting and Boosting for Few-Shot Segmentation},
  booktitle = ICCV,
  year      = {2019},
}

@inproceedings{gupta2019lvis,
  author    = {Agrim Gupta and Piotr Doll{\'a}r and Ross Girshick},
  title     = {{LVIS}: A Dataset for Large Vocabulary Instance Segmentation},
  booktitle = CVPR,
  year      = {2019}
}

@misc{google2025nanobananapro,
  author       = {{Google DeepMind}},
  title        = {Introducing {Nano Banana Pro}: {Gemini}~3 {Pro} {Image}},
  year         = {2025},
  howpublished = {Google blog post},
  url          = {https://blog.google/technology/ai/nano-banana-pro/},
}

@misc{openai2025gptimage,
  author = {OpenAI},
  title = {{The new ChatGPT Images is here}},
  year = {2025},
  howpublished = {\url{https://openai.com/index/new-chatgpt-images-is-here/}},
}

@article{bytedance2025seedream4,
  author  = {{ByteDance Seed Team}},
  title   = {{Seedream}~4.0: Toward Next-generation Multimodal Image Generation},
  journal = {arXiv preprint arXiv:2509.20427},
  year    = {2025},
}

@misc{black2025flux2,
  author = {Black Forest Labs},
  title = {{FLUX.2: Frontier Visual Intelligence}},
  year = {2025},
  howpublished = {\url{https://bfl.ai/blog/flux-2}},
}

@misc{google2025gemini3,
  author       = {{Google DeepMind}},
  title        = {{Gemini}~3 {Flash}},
  year         = {2025},
  howpublished = {Google DeepMind model page},
  url          = {https://deepmind.google/models/gemini/flash/},
}

@article{li2026openvton,
  title   = {{OpenVTON-Bench}: A Large-Scale High-Resolution Benchmark for Controllable Virtual Try-On Evaluation},
  author  = {Li, Jin and Chen, Tao and Jiang, Shuai and Wang, Weijie and Luo, Jingwen and Wu, Chenhui},
  journal = {arXiv preprint arXiv:2601.22725},
  year    = {2026}
}

@article{chen2026tstars,
  title   = {{Tstars-Tryon} 1.0: Robust and Realistic Virtual Try-On for Diverse Fashion Items},
  author  = {Chen, Mengting and Chen, Zhengrui and Du, Yongchao and Gao, Zuan and Hu, Taihang and Lan, Jinsong and Lin, Chao and Shen, Yefeng and Wang, Xingjian and Wang, Zhao and others},
  journal = {arXiv preprint arXiv:2604.19748},
  year    = {2026}
}

@inproceedings{guler2018densepose,
  author    = {R{\i}za Alp G{\"u}ler and Natalia Neverova and Iasonas Kokkinos},
  title     = {{DensePose}: Dense Human Pose Estimation in the Wild},
  booktitle = CVPR,
  year      = {2018}
}

@inproceedings{khirodkar2024sapiens,
  author    = {Rawal Khirodkar and Timur Bagautdinov and Julieta Martinez and Su Zhaoen and Austin James and Peter Selednik and Stuart Anderson and Shunsuke Saito},
  title     = {{Sapiens}: Foundation for Human Vision Models},
  booktitle = ECCV,
  year      = {2024}
}

@inproceedings{he2016resnet,
  author    = {Kaiming He and Xiangyu Zhang and Shaoqing Ren and Jian Sun},
  title     = {Deep Residual Learning for Image Recognition},
  booktitle = CVPR,
  year      = {2016}
}

@article{oquab2024dinov2,
  author  = {Maxime Oquab and Timoth{\'e}e Darcet and Th{\'e}o Moutakanni and Huy V. Vo and Marc Szafraniec and Vasil Khalidov and Pierre Fernandez and Daniel Haziza and Francisco Massa and Alaaeldin El-Nouby and Mahmoud Assran and Nicolas Ballas and Wojciech Galuba and Russell Howes and Po-Yao Huang and Shang-Wen Li and Ishan Misra and Michael Rabbat and Vasu Sharma and Gabriel Synnaeve and Hu Xu and Herv{\'e} J{\'e}gou and Julien Mairal and Patrick Labatut and Armand Joulin and Piotr Bojanowski},
  title   = {{DINOv2}: Learning Robust Visual Features without Supervision},
  journal = TMLR,
  year    = {2024},
}

@article{simeoni2025dinov3,
  author  = {Oriane Sim{\'e}oni and Huy V. Vo and Maximilian Seitzer and Federico Baldassarre and Maxime Oquab and Cijo Jose and Vasil Khalidov and Marc Szafraniec and Seungeun Yi and Micha{\"e}l Ramamonjisoa and Francisco Massa and Daniel Haziza and Luca Wehrstedt and Jianyuan Wang and Timoth{\'e}e Darcet and Th{\'e}o Moutakanni and Leonel Sentana and Claire Roberts and Andrea Vedaldi and Jamie Tolan and John Brandt and Camille Couprie and Julien Mairal and Herv{\'e} J{\'e}gou and Patrick Labatut and Piotr Bojanowski},
  title   = {{DINOv3}},
  journal = {arXiv preprint arXiv:2508.10104},
  year    = {2025}
}

@inproceedings{li2023gligen,
  author    = {Yuheng Li and Haotian Liu and Qingyang Wu and Fangzhou Mu and Jianwei Yang and Jianfeng Gao and Chunyuan Li and Yong Jae Lee},
  title     = {{GLIGEN}: Open-Set Grounded Text-to-Image Generation},
  booktitle = CVPR,
  year      = {2023}
}

@inproceedings{wang2024instancediffusion,
  author    = {Xudong Wang and Trevor Darrell and Sai Saketh Rambhatla and Rohit Girdhar and Ishan Misra},
  title     = {{InstanceDiffusion}: Instance-Level Control for Image Generation},
  booktitle = CVPR,
  year      = {2024}
}

@inproceedings{wang2024msdiffusion,
  author    = {Xierui Wang and Siming Fu and Qihan Huang and Wanggui He and Hao Jiang},
  title     = {{MS-Diffusion}: Multi-Subject Zero-Shot Image Personalization with Layout Guidance},
  booktitle = ICLR,
  year      = {2025}
}

@inproceedings{choi2021vitonhd,
  author    = {Seunghwan Choi and Sunghyun Park and Minsoo Kang and Jaegul Choo},
  title     = {{VITON-HD}: High-Resolution Virtual Try-On via Misalignment-Aware Normalization},
  booktitle = CVPR,
  year      = {2021}
}

@inproceedings{feng2025omnitry,
  author    = {Yutong Feng and Linlin Zhang and Hengyuan Cao and Yiming Chen and Xiaoduan Feng and Jian Cao and Yuxiong Wu and Bin Wang},
  title     = {{OmniTry}: Virtual Try-On Anything without Masks},
  booktitle = NeurIPS,
  year      = {2025}
}

@article{hu2026garments2look,
  title={Garments2Look: A Multi-Reference Dataset for High-Fidelity Outfit-Level Virtual Try-On with Clothing and Accessories},
  author={Hu, Junyao and Cheng, Zhongwei and Wong, Waikeung and Zou, Xingxing},
  journal={arXiv preprint arXiv:2603.14153},
  year={2026}
}

@inproceedings{heusel2017fid,
  author    = {Martin Heusel and Hubert Ramsauer and Thomas Unterthiner and Bernhard Nessler and Sepp Hochreiter},
  title     = {{GANs} Trained by a Two Time-Scale Update Rule Converge to a Local Nash Equilibrium},
  booktitle = NeurIPS,
  year      = {2017}
}

@article{wang2004ssim,
  author    = {Zhou Wang and Alan C. Bovik and Hamid R. Sheikh and Eero P. Simoncelli},
  title     = {Image Quality Assessment: From Error Visibility to Structural Similarity},
  journal   = {IEEE Transactions on Image Processing},
  volume    = {13},
  number    = {4},
  pages     = {600--612},
  year      = {2004}
}

@inproceedings{zhang2018lpips,
  author    = {Richard Zhang and Phillip Isola and Alexei A. Efros and Eli Shechtman and Oliver Wang},
  title     = {The Unreasonable Effectiveness of Deep Features as a Perceptual Metric},
  booktitle = CVPR,
  year      = {2018}
}

@inproceedings{caron2021dino,
  author    = {Mathilde Caron and Hugo Touvron and Ishan Misra and Herv{\'e} J{\'e}gou and Julien Mairal and Piotr Bojanowski and Armand Joulin},
  title     = {Emerging Properties in Self-Supervised Vision Transformers},
  booktitle = ICCV,
  year      = {2021}
}

@inproceedings{radford2021clip,
  author    = {Alec Radford and Jong Wook Kim and Chris Hallacy and Aditya Ramesh and Gabriel Goh and Sandhini Agarwal and Girish Sastry and Amanda Askell and Pamela Mishkin and Jack Clark and Gretchen Krueger and Ilya Sutskever},
  title     = {Learning Transferable Visual Models From Natural Language Supervision},
  booktitle = ICML,
  year      = {2021}
}

@article{kim2026coral,
  title={CORAL: Correspondence Alignment for Improved Virtual Try-On},
  author={Kim, Jiyoung and Shin, Youngjin and Jin, Siyoon and Chung, Dahyun and Nam, Jisu and Kim, Tongmin and Park, Jongjae and Kang, Hyeonwoo and Kim, Seungryong},
  journal={arXiv preprint arXiv:2602.17636},
  year={2026}
}

@article{huang2024context,
  title={In-context lora for diffusion transformers},
  author={Huang, Lianghua and Wang, Wei and Wu, Zhi-Fan and Shi, Yupeng and Dou, Huanzhang and Liang, Chen and Feng, Yutong and Liu, Yu and Zhou, Jingren},
  journal={arXiv preprint arXiv:2410.23775},
  year={2024}
}

@article{hu1962moments,
  author  = {Ming-Kuei Hu},
  title   = {Visual Pattern Recognition by Moment Invariants},
  journal = {IRE Trans. Inf. Theory},
  volume  = {8},
  number  = {2},
  pages   = {179--187},
  year    = {1962}
}

@article{huttenlocher1993hausdorff,
  author  = {Daniel P. Huttenlocher and Gregory A. Klanderman and William J. Rucklidge},
  title   = {Comparing Images Using the {Hausdorff} Distance},
  journal = PAMI,
  volume  = {15},
  number  = {9},
  pages   = {850--863},
  year    = {1993}
}

@article{yang2025fitcontroler,
  author  = {Lu Yang and Yicheng Liu and Yanan Li and Xiang Bai and Huchuan Lu},
  title   = {{FitControler}: Toward Fit-Aware Virtual Try-On},
  journal = {arXiv preprint arXiv:2512.24016},
  year    = {2025}
}

@article{zheng2025lanpaint,
  author  = {Candi Zheng and Yuan Lan and Yang Wang},
  title   = {LanPaint: Training-Free Diffusion Inpainting with Asymptotically Exact and Fast Conditional Sampling},
  journal = TMLR,
  year    = {2025}
}

@inproceedings{levin2025differential,
  title={Differential diffusion: Giving each pixel its strength},
  author={Levin, Eran and Fried, Ohad},
  booktitle={Computer Graphics Forum},
  volume={44},
  number={2},
  year={2025},
  organization={Wiley Online Library}
}

\newpage
\appendix
\renewcommand{\theHsection}{appendix.\thesection}
\renewcommand{\theHsubsection}{appendix.\thesubsection}
\renewcommand{\theHsubsubsection}{appendix.\thesubsubsection}

\section*{Supplementary Material}
\addcontentsline{toc}{section}{Supplementary Material}

\section{VIP-SAM: Resource Comparison}
\label{sec:supp_vipsam_resources}

We complement the segmentation accuracy results in Tab.~\ref{tab:comparison_miou} of the main paper with a comparison of the resource footprint of each method.
Tab.~\ref{tab:comparison_resources} shows peak training memory, peak inference memory, FLOPs, total parameter count, and the number of trainable parameters.
Three observations stand out.
First, VIP-SAM (ViT-B/ResNet-50) requires substantially higher training resource than VRP-SAM (ViT-B/ResNet-50) because gradient must be computed for the layers inside the ViT backbone, but the inference resource is comparable.
From architectural point of view, these two configurations are directly comparable, and the results in Tab.~\ref{tab:comparison_miou} of the main text support the validity of our design choices.
Second, VIP-SAM (especially Hiera variants) achieves higher accuracy with a substantially smaller resource footprint (memory, FLOPs, parameters) than the SAM-ViT-H-based baselines: even our largest variant (Hiera-L + DINOv3-H) has inference memory comparable to PerSAM/Matcher while reaching state-of-the-art accuracy.
Third, despite training more parameters than VRP-SAM/ProSAM, our smaller variants use less inference memory and fewer FLOPs, owing to the lighter Hiera backbones.

\begin{table}[h]
\centering
\caption{Resource comparison for visual-reference segmentation methods. ``--'' denotes metrics that are not applicable: PerSAM and Matcher are training-free.}
\label{tab:comparison_resources}
\setlength{\tabcolsep}{4pt}
\resizebox{\linewidth}{!}{%
\begin{tabular}{lccccc}
\toprule
Method & Train Mem (GiB) & Inference Mem (GiB) & FLOPs ($\times 10^{12}$) & Tot Params (M) & Train Params (M) \\
\midrule
PerSAM~\cite{zhang2023personalize} (ViT-H / ViT-H)  & --   & 6.47 & 11.9 & 641   & --   \\
Matcher~\cite{liu2023matcher} (ViT-H / DINOv2-L)    & --   & 7.65 & 8.07 & 945   & --   \\
VRP-SAM~\cite{sun2024vrpsam} (ViT-B/ResNet-50)      & 14.2 & 4.07 & 1.48 & 118   & 1.59 \\
VRP-SAM~\cite{sun2024vrpsam} (ViT-H/ResNet-50)      & 18.5 & 7.05 & 6.47 & 666   & 1.59 \\
ProSAM~\cite{wang2025prosam} (ViT-H/ResNet-50)      & 19.1 & 7.05 & 6.47 & 666   & 1.72 \\
\midrule
\rowcolor{ourRow}
VIP-SAM (ViT-B / ResNet-50)          & 42.4 & 4.87 & 3.77 & 161     & 49.0 \\
\rowcolor{ourRow}
VIP-SAM (Hiera-B+ / ResNet-50)       & 20.1 & 2.78 & 1.43 & 128     & 35.9 \\
\rowcolor{ourRow}
VIP-SAM (Hiera-L / ResNet-50)        & 30.5 & 3.30 & 2.63 & 274     & 38.5 \\
\rowcolor{ourRow}
VIP-SAM (Hiera-L / DINOv3-H)         & 32.2 & 6.50 & 4.43 & 1{,}090 & 37.4 \\
\bottomrule
\end{tabular}%
}
\end{table}

\section{Inpainting Failure Modes: Visual Examples}
\label{sec:supp_inpainting_failures}

In {Sec.~\ref{sec:intro}} and {Sec.~\ref{sec:cvton_formulation}} of the main paper, we claimed that inpainting-based VTO is fundamentally limited by the mask itself.
We justify this claim through Fig.~\ref{fig:supp_inpainting_failures}, which shows two examples and four annotated failure regions.
The first three regions show that the generated output is highly sensitive to the \emph{size and shape} of the mask; the last shows that the output is also entangled with the \emph{context outside} the mask.

\begin{figure}[h]
\centering
\includegraphics[width=\linewidth]{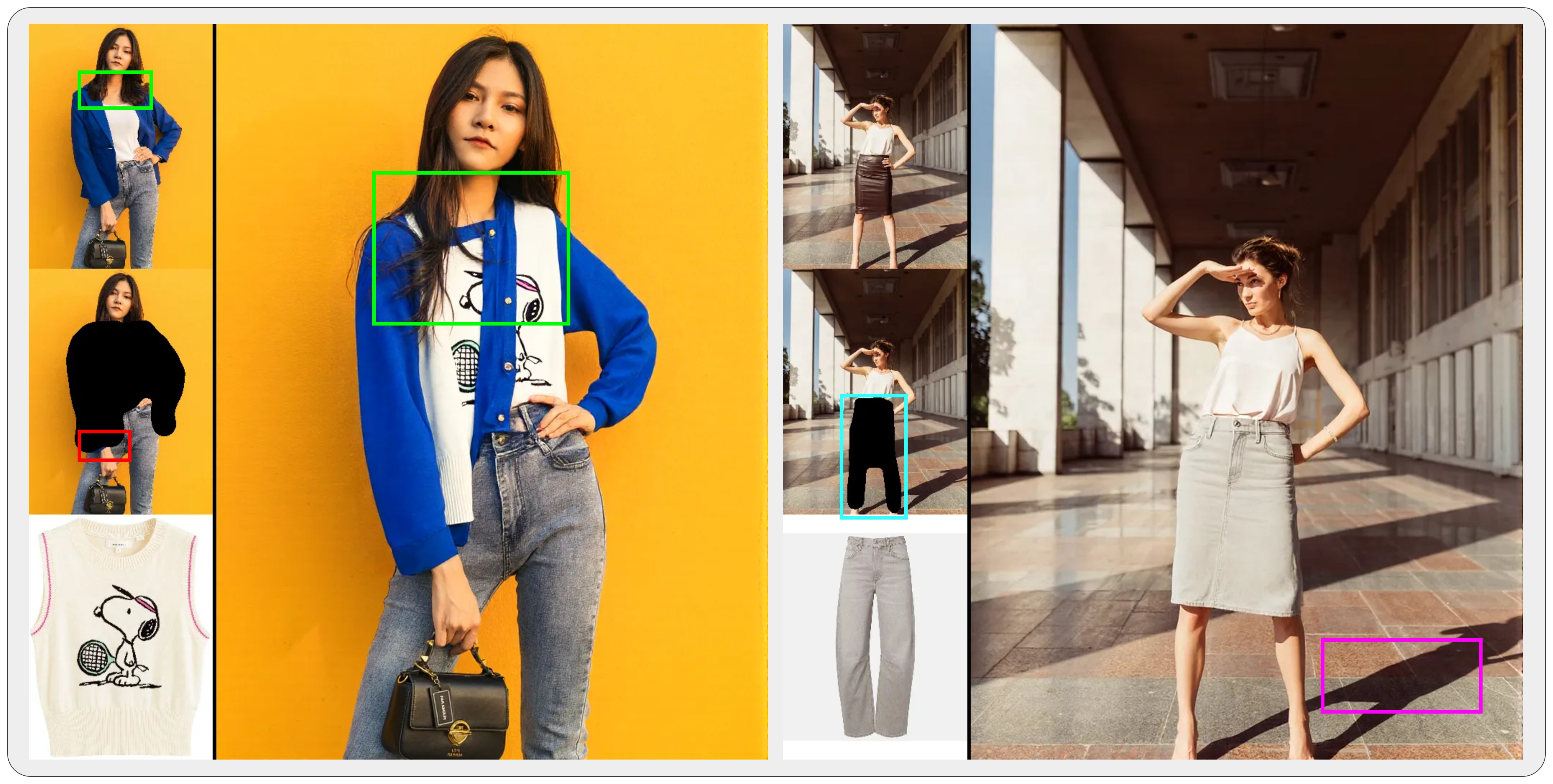}
\caption{\textbf{Failure modes of inpainting-based VTO.} Each example shows the input person, the masked person fed to the inpainting model, the reference garment, and the inpainted output.
\textbf{\textcolor{red}{Red}}: an undersized mask leaves residual pixels of the original garment (blue jacket sleeve); the inpainting model treats them as context and renders the new garment incorrectly.
\textbf{\textcolor{green}{Green}}: enlarging the mask to avoid the above issue erases identity information (pose, face, hair, skin and tattoos, held items like bags or phones), which the inpainting model must then hallucinate (cf.~Fig.~\ref{fig:vto_qual}, rows 2--3).
\textbf{\textcolor{cyan}{Cyan}}: the output is biased by the \emph{shape} of the mask itself. A trapezoidal erasure region cues the inpainting model to synthesize a skirt even though the reference shows pants.
\textbf{\textcolor{magenta}{Magenta}}: the output is also biased by context \emph{outside} the mask. The shadow of the original skirt on the floor remains visible and the inpainting model harmonizes the new garment with that shadow, again producing a skirt rather than the reference pants.
These examples illustrate why mask-completion is inadequate for VTO: it is heavily influenced by mask quality and the context, both of which are difficult to control.}
\label{fig:supp_inpainting_failures}
\end{figure}

\section{Data Pipeline Details}
\label{sec:supp_data}

\subsection{Data Sources and Proportions}
\label{sec:supp_data_sources}
The person--garment pairs used to train CtrlVTON are drawn from three complementary sources: (i) publicly available VTO datasets, (ii) commercial datasets licensed from fashion retailers, and (iii) in-house datasets collected specifically for this work.
A coarse breakdown of source-level proportions and garment-category coverage is reported in Table~\ref{tab:supp_data_breakdown}.

\begin{table}[h]
\centering
\caption{Coarse composition of the CtrlVTON training corpus.}
\label{tab:supp_data_breakdown}
\begin{tabular}{lc}
\toprule
Source & Share \\
\midrule
Public VTO datasets                & 32\% \\
Licensed commercial datasets       & 62\% \\
In-house datasets                  & 6\% \\
\bottomrule
\end{tabular}
\end{table}

We further break down the corpus along three axes that are directly relevant to the capabilities evaluated in {Sec.~\ref{sec:eval}} of the main paper: the single- vs.\ multi-garment split, the distribution over garment categories, and the distribution over task tokens ({Sec.~\ref{sec:cvton_model}} of the main paper).
Table~\ref{tab:supp_data_composition} reports these statistics.
A multi-garment sample is labeled \emph{mixed} in the per-sample task-token breakdown when its constituent garments are assigned more than one distinct task token (e.g., one garment tagged \textsc{full\_swap} while another in the same sample is tagged \textsc{add}); this case has no analogue in the per-garment breakdown, where every garment carries exactly one task token by construction.

\begin{table}[h]
\centering
\caption{Composition of the CtrlVTON training corpus by garment cardinality, garment category, and task-token assignment.}
\label{tab:supp_data_composition}
\begin{tabular}{@{}l@{\hspace{10pt}}|@{\hspace{10pt}}l@{}}
\toprule
\begin{tabular}[t]{ll}
\multicolumn{2}{l}{\textit{Garment cardinality (per sample)}} \\
Single-garment & 51\% \\
Multi-garment  & 49\% \\
\addlinespace
\multicolumn{2}{l}{\textit{Garment category (per garment)}} \\
Upper & 41\% \\
Lower & 24\% \\
Full  & 18\% \\
Bag   & 7\%  \\
Shoes & 10\% \\
\end{tabular}
&
\begin{tabular}[t]{ll}
\multicolumn{2}{l}{\textit{Task token (per garment)}} \\
\textsc{full\_swap}    & 68\% \\
\textsc{partial\_swap} & 18\% \\
\textsc{add}           & 14\% \\
\addlinespace
\multicolumn{2}{l}{\textit{Task token (per sample)}} \\
\textsc{full\_swap}    & 70\% \\
\textsc{partial\_swap} & 11\% \\
\textsc{add}           & 7\%  \\
Mixed                  & 12\% \\
\end{tabular}
\\
\bottomrule
\end{tabular}
\end{table}

\subsection{Masking Strategies for Synthetic $p_{\text{ref}}$}
\label{sec:supp_masking}

To obtain synthetic images for our training data, we utilize inpainting models~\cite{flux2024,lee2025voost} and editing models~\cite{labs2025flux1kontextflowmatching,black2025flux2} with training-free inpainting methods~\cite{levin2025differential,zheng2025lanpaint}.
We will refer to both methods as an inpainting system for convenience.

The mask supplied to the inpainting system controls how much of the original image is allowed to change.
A fixed strategy is insufficient because each image requires different trade-offs between context preservation, garment-shape variation, and localization.
We therefore use three masking strategies, summarized below in order from our most frequently used (\emph{default}) to the most surgical (\emph{exception}).

\noindent\textbf{Box mask (default).}
Used as the default for the majority of the corpus.
Starting from a \emph{garment-agnostic} body region produced by off-the-shelf human segmentation models~\cite{guler2018densepose,khirodkar2024sapiens}, we extract the axis-aligned bounding box of the garment region, and exclude face, hair, hands, and held items.
The box's extent is independent of the original garment's silhouette, so the inpainting system has the freedom to render the new garment in a wide range of shapes, producing the highest \emph{garment-shape diversity} in the synthetic data.

\noindent\textbf{Loose mask.}
Used when free-form variation is unsafe, \eg, the background contains complex structure that should not change, or the person is holding accessories (bags, jewelry) that must be preserved.
We again build a garment-agnostic mask using human segmentation models~\cite{guler2018densepose,khirodkar2024sapiens}, but keep its segmentation contour rather than converting it to a bounding box.
This confines the edit to the body silhouette while allowing natural garment-shape variation.

\noindent\textbf{Tight mask.}
Used to synthesize data for operations that demand precise spatial control---most importantly the \textsc{partial\_swap} task token ({Sec.~\ref{sec:cvton_model}} of the main paper), where only a single specific garment among several should be modified.
The tight mask is the per-instance VIP-SAM mask of the original garment, dilated by a small margin.
This concentrates the edit in the exact region intended for change and leaves the rest of the body untouched.

We generate at least one candidate per strategy and use the VLM + human screening described below to pick the best $p_{\text{ref}}$ for each image.

\subsection{Quality-Control Protocol}
\label{sec:supp_qc}

For every source pair $(p, g_{\text{ref}})$ we synthesize four to five candidate $p_{\text{ref}}$ images using inpainting systems with different masking strategies.
The candidates then pass through a three-stage funnel: a VLM-based screen, an automatic silhouette-leakage filter, and a final human review that selects the  best candidate.

\noindent\textbf{Stage 1: VLM-based screen.}
A vision--language model is prompted to answer four yes/no questions comparing each candidate $p_{\text{ref}}$ to $p$:
(1) Is the person identity preserved? (2) Is the pose preserved? (3) Is the background preserved? (4) Is the modification confined to a garment region?
A candidate must receive ``yes'' on all four questions to proceed to the next stage.

\noindent\textbf{Stage 2: Silhouette-leakage filter.}
When fine-tuning a pre-trained editing model for VTO, it tends to preserve the silhouette of $p_{\text{ref}}$.
Therefore, training on pairs $(p, p_{\text{ref}})$ with near-identical silhouettes reinforces the model's tendency to ignore the shape of $g_{\text{ref}}$ (Fig.~\ref{fig:supp_qual_silhouette}).
In our case, since both VIP-SAM masks $M_p$ and $M_{p_{\text{ref}}}$ are easily obtained, we use them directly in the screening process.
Let $A$ and $B$ denote the two binary masks, and let $\partial A,\partial B$ be their $1$-pixel-thick boundary point sets.
Given the tolerance $\tau$ px we define the directional contour match fractions
\begin{equation}
f_{A \to B} = \frac{\bigl|\{\,p \in \partial A \;:\; d(p, \partial B) \le \tau\,\}\bigr|}{|\partial A|},
\qquad
f_{B \to A} = \frac{\bigl|\{\,p \in \partial B \;:\; d(p, \partial A) \le \tau\,\}\bigr|}{|\partial B|},
\end{equation}
where $d(p, S) = \min_{q \in S} \lVert p - q \rVert_2$ is the Euclidean distance from $p$ to the closest point in $S$, computed efficiently via a distance transform.
We then define the \emph{contour match fraction} as
\begin{equation}
\mathrm{CMF}(A, B) = \max\ \!\bigl(f_{A \to B},\; f_{B \to A}\bigr),
\end{equation}
and discard any candidate whose $\mathrm{CMF}(M_p, M_{p_{\text{ref}}})$ exceeds a threshold.
Intuitively, $\mathrm{CMF}$ is the largest fraction of one mask's contour that lies within $\tau$ pixels of the other mask's contour; a high value means the edited and original garments share most of their silhouette.
We use this contour-based score rather than mask IoU because in the case of partial silhouette overlap (\eg, identical sleeves and hem with a different neckline), IoU remains only moderate even though most of the contour is leaking, see Fig.~\ref{fig:supp_silhouette_filter} (d).
The maximum in CMF makes the score symmetric and robust to one mask being a strict subset of the other.
$\mathrm{CMF}$ is also distinct from the symmetric Hausdorff distance $d_H$ used in our mask-adherence metrics (Appendix~\ref{sec:supp_mask_metrics}): $d_H$ captures the \emph{worst-case} pointwise deviation between two contours, whereas $\mathrm{CMF}$ captures the \emph{bulk} fraction of contour that is within the tolerance, which is what we need to detect silhouette leakage.

\begin{figure}[h]
\centering
\includegraphics[width=\linewidth]{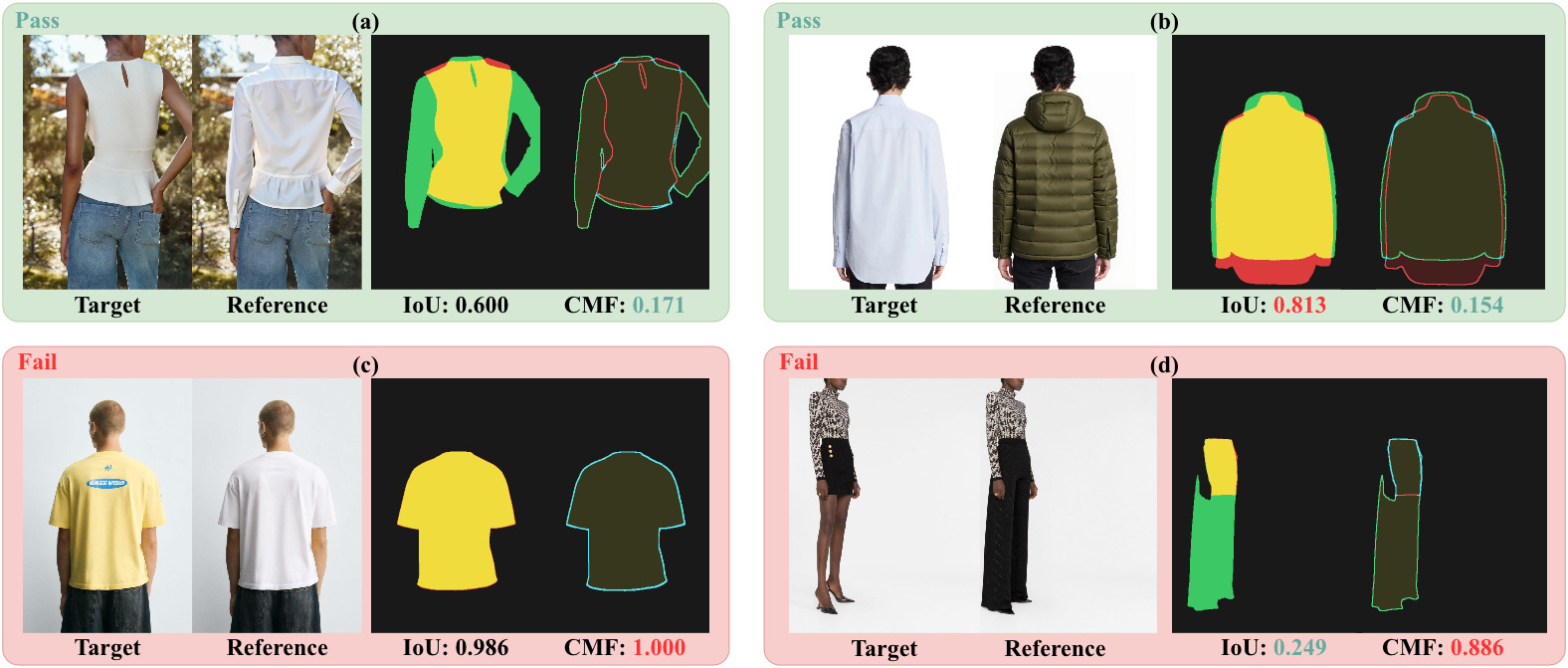}
\caption{\textbf{Examples illustrating the contour match fraction (CMF) filter and its advantage over IoU.} Each row shows a $(p, p_{\text{ref}})$ pair together with the two VIP-SAM masks $M_p$ and $M_{p_{\text{ref}}}$, and reports both $\mathrm{IoU}(M_p, M_{p_{\text{ref}}})$ and $\mathrm{CMF}(M_p, M_{p_{\text{ref}}})$. Pairs marked \textit{fail} (high $\mathrm{CMF}$) are discarded because their silhouettes are nearly identical: training on them would teach the model to ignore $g_{\text{ref}}$ and copy the reference contour. Case~(d) shows an example that an IoU-based filter would miss.}
\label{fig:supp_silhouette_filter}
\end{figure}

\noindent\textbf{Stage 3: Human review.}

Three annotators inspect every candidate that survives Stages 1 and 2.
Their job is twofold: catch failure cases the automatic filters missed (\eg, subtle identity drift, shadow inconsistency, implausible garment proportions) and select the best candidate per source pair by overall physical plausibility.
The top candidate becomes the $p_{\text{ref}}$ used for that training instance; all others are discarded.

\section{Training Details}
\label{sec:supp_training}

\noindent{\textbf{Visual-Instance-Prompt Segmentation}}
Our training configuration mostly follows VRP-SAM's since our innovation is in the architecture.
We randomly split the fashion dataset into train, val, and test sets according to the ratio $85 : 7.5 : 7.5$.
On this dataset, we use the same lr scheduler and loss as VRP-SAM, but change batch size to 16 and train for 100 epochs.
Note that 16 is the maximum batch size that fits on an L40 GPU for ViT-B / ResNet-50 configuration.
For fairness, we use the same configuration when training VRP-SAM and ProSAM on this dataset.

When training on the COCO-$20^i$ and PASCAL-$5^i$ datasets, we use the splits (i.e. the partition of classes into training and held-out) provided by VRP-SAM and report the average across splits.
Unlike standard few-shot segmentation, we evaluate on the same classes used during training rather than the held-out classes, so that the evaluation becomes similar to our task---finding the same instance---rather than generalization to novel classes.
We train all models (VIP-SAM, VRP-SAM, ProSAM) for 50 epochs, matching VRP-SAM's original configuration.

Interestingly, ProSAM outperforms VRP-SAM on the PASCAL-$5^i$ dataset but not on the COCO-$20^i$ dataset.
Upon closer examination, we find that images in COCO-$20^i$ often contain multiple same-category instances of varying size and shape.
A simple proxy for this is the number of connected components per annotation mask (this can both overestimate and underestimate the object count: adjacent objects merge into a single component, while individual objects can fragment into several).
After filtering small mask components, which are usually just noise, we find that on average, a mask in COCO-$20^i$ contains 1.91 components per mask vs 1.77 in PASCAL-$5^i$.
This is telling since the trick introduced in ProSAM pushes the support image embedding into a flat region of the loss landscape, which acts like a regularizer.
When the support image contains multiple visually distinct objects of the same class, the resulting embedding becomes an "average" over them.
While this regularization helps generalization to novel classes, it can hurt when evaluation classes match training classes.
This is especially true when the support image contains multiple instances: while ProSAM is forced to average over them, an unregularized embedding could instead collapse onto a single instance, which is what instance-level identification needs.

\noindent\textbf{CtrlVTON-base}
CtrlVTON-base is obtained by full-parameter fine-tuning of FLUX.2 Klein~\cite{labs2025flux1kontextflowmatching} on the triplets $(p_{\text{ref}}, g_{\text{ref}}, p)$.
Training runs at $\sim$1MP input resolution with a global batch size of 128.
The total compute budget is approximately 20 H200-days (i.e., 20 NVIDIA H200 GPUs for approximately one day of continuous training).

\section{Mask Injection Strategies: Channel-Wise vs.\ Token-Wise}
\label{sec:supp_injection}

In {Sec.~\ref{sec:cvton_model}} of the main paper, we concatenated the mask tokens \emph{channel-wise} with their corresponding image tokens. A natural alternative is to concatenate them along the \emph{token} dimension instead, treating the mask tokens as additional reference tokens as in IC-LoRA~\cite{huang2024context} and CORAL~\cite{kim2026coral}. Here we justify our choice of channel-wise over token-wise concatenation.

\noindent\textbf{Design rationale.}
Channel-wise concatenation has two properties that are desirable in our setting.
First, every mask is spatially aligned with its corresponding image by construction: $M_p$ shares the same pixel grid as $p$, $M_{p_{\text{ref}}}$ as $p_{\text{ref}}$, and $M_{g_{\text{ref}}}$ as $g_{\text{ref}}$.
Concatenating along the channel dimension preserves this alignment for free.
Token-wise concatenation instead treats each mask as an additional reference block, forcing the network to re-discover the spatial correspondence through attention.
Second, the cost of self-attention is quadratic in the number of tokens, so concatenating three full-resolution mask tokens inflates both compute and memory at every attention layer, which is particularly inefficient for high resolution images.

\noindent\textbf{Empirical comparison.}
We compare the two concatenation strategies under the same training budget on the VITON-HD-edit benchmark, and report trainable parameter count, inference throughput, peak inference VRAM, and mask-adherence and perceptual metrics.

\begin{table}[h]
\centering
\caption{Channel-wise vs.\ token-wise mask injection. The two variants start from the same base checkpoint and are trained for the same number of steps.}
\label{tab:supp_injection}
\setlength{\tabcolsep}{4pt}
\resizebox{\linewidth}{!}{%
\begin{tabular}{lccccccccc}
\toprule
Injection scheme & Trainable params (M) & Inference speed (it/s) & VRAM (GB)
 & IoU$\uparrow$ & $d_{\text{Hu}}\downarrow$ & $d_H\downarrow$ & M-DINO$\uparrow$ & M-CLIP-I$\uparrow$ & GTC$\uparrow$ \\
\midrule
Token-wise & 572 & 0.38 & 62.65 & 0.8925 & 0.0034 & 31.84 & 0.7821 & 0.8987 & 4.1985 \\
\rowcolor{ourRow}
Channel-wise & 572 & 0.91 & 43.47 & 0.9610 & 0.0022 & 26.05 & 0.8212 & 0.9052 & 4.2773 \\
\bottomrule
\end{tabular}%
}
\end{table}

\section{Evaluation Metrics: Details}
\label{sec:supp_metrics}
In this section, we present additional details on garment-fidelity, VLM-as-judge, and mask-adherence, which were summarized in {Sec.~\ref{sec:exp_setup}} of the main paper\,(a)--(c).

\subsection{Garment-fidelity Metrics (M-DINO, M-CLIP-I)}
\label{sec:supp_metrics_object}
We adopt the garment-fidelity metrics of OmniTry~\cite{feng2025omnitry}, which are designed for the mask-free editing setting where no per-instance ground-truth image is available.
For each generated try-on image we crop the garment region using its VIP-SAM mask ({Sec.~\ref{sec:vipseg}} of the main paper), apply white-background normalization to remove surrounding context, and compare the result with an identically processed reference garment in two embedding spaces.
For each encoder we report the cosine similarity, which lies in $[-1,1]$, higher being better:
\begin{itemize}
\item \textbf{M-DINO}: features from a self-supervised ViT trained with DINO~\cite{caron2021dino} capture fine-grained local structure, so this score is sensitive to structural details of garment parts.
\item \textbf{M-CLIP-I}: features from the CLIP~\cite{radford2021clip} image encoder are aligned with semantic concepts, so this score primarily reflects category-level coherence.
\end{itemize}
Empirically, M-DINO scores tend to be lower than M-CLIP-I on the same outputs because the DINO embedding penalizes geometric variation that the CLIP embedding is largely invariant to.
The two metrics are therefore most informative when read together: a method that scores well on M-CLIP-I but poorly on M-DINO preserves the garment category but not its precise local structure, whereas a method strong on both is faithful at both the category and structural levels.

\subsection{VLM-as-Judge: GTC / PBC / PR}
\label{sec:supp_metrics_vlm}
We use Gemini 3.0 Flash~\cite{google2025gemini3} as a judge to score try-on outputs along three VTO quality criteria.
Recent work has shown that traditional distributional metrics such as FID inadequately capture VTO-specific quality, motivating VLM-based evaluation protocols~\cite{kim2026coral,li2026openvton,chen2026tstars}.
CORAL~\cite{kim2026coral} introduces Garment Transfer Consistency (GTC) and Fit Pose Coherence (FPC) to separately assess garment fidelity and wearing plausibility.
OpenVTON-bench~\cite{li2026openvton} decomposes try-on quality into five criteria (background, identity, texture, shape, realism), arguing that a single aggregate score hides failures specific to individual criteria.
Tstars-Tryon 1.0~\cite{chen2026tstars} organizes the criteria into two stages: a \emph{garment-aware} stage that evaluates identity consistency and garment fidelity given the reference garment as context, and a \emph{garment-agnostic} stage that evaluates background preservation and physical realism without the reference garment, isolating these aspects from garment-induced bias.
Following the broader VLM-as-judge literature, we adopt three criteria to evaluate VTO quality:
\begin{itemize}
\item \textbf{GTC} (Garment Transfer Consistency): how faithfully the reference garment is reproduced on the person, from \emph{local details} (prints, buttons, zippers, pockets) to \emph{global properties} (texture, color, silhouette).
\item \textbf{PBC} (Person-Background Consistency): how well the input person and surroundings are preserved, including identity (face, hair, skin, tattoos), pose, body shape, personal items (bags, phones, watches, jewelry), and background.
\item \textbf{PR} (Physical Realism): whether the garment is worn in a physically plausible manner, including drape and folds, contact with the body, consistent occlusion with other garments and accessories, and lighting/shadow consistency.
\end{itemize}
Each metric is scored on a $[0, 5]$ float scale, where 5 means ``no perceivable issues'' and 0 means ``severe failure on every aspect.''
The grader is instructed to evaluate \emph{strictly}: every visible artifact, mismatch, or implausibility deducts points, and a perfect score is reserved for outputs that withstand close scrutiny.
The exact prompt we send to the model is below.

\FloatBarrier
\begin{tcolorbox}[
  colback=gray!5, colframe=gray!50!black,
  title={VLM-as-judge prompt (verbatim)},
  fonttitle=\bfseries\small,
  fontupper=\scriptsize,
  boxrule=0.4pt, arc=1pt, left=4pt, right=4pt, top=4pt, bottom=4pt,
  breakable
]
\begin{verbatim}
You are a strict evaluator of virtual try-on results. You will be
shown the following images:

  [INPUT_PERSON]      - the original photograph of the person.
  [REFERENCE_GARMENT] - the garment(s) that the person should be wearing
                        (could be a single item or a set of multiple items).
  [GENERATED_IMAGE]   - the model's output.

Score the GENERATED_IMAGE along three orthogonal axes. Each score is
a float in [0, 5]. Be STRICT: 5 is reserved for outputs with no
perceivable issues; deduct points for any visible artifact, mismatch, 
or implausibility.

  GTC (Garment Transfer Consistency, [0, 5]):
    Does the rendered garment match the REFERENCE_GARMENT in BOTH
    fine details (prints, buttons, zippers, logos, text) AND global 
    properties (texture, color, silhouette)?
    - (Multi-garment): Check if ALL provided reference garments are 
      present and individually accurate. Deduct points if any item 
      is missing, simplified, or shows color bleeding from other items.

  PBC (Person-Background Consistency, [0, 5]):
    Are the input person and surroundings preserved by the edit?
    - Identity & Body: Face, hair, skin tone, tattoos, and proportions.
    - Non-target Areas: Preservation of held belongings (bags, phones) 
      and original garments NOT meant to be changed (e.g., shoes, hats).
    - Background: Scene integrity, lighting, and lack of warping.

  PR (Physical Realism, [0, 5]):
    Is the garment worn in a physically plausible way?
    - Drape and folds consistent with gravity and body contour.
    - Correct contact: No floating fabric or clipping through skin.
    - (Multi-garment): Check the interaction between items. 
      Outerwear must correctly cover inner layers; tops should be 
      naturally tucked in or layered over bottoms without unrealistic 
      merging or clipping.
    - Lighting and shadows on the new garment(s) match the scene.

Return your answer as a single JSON object, no extra text:

  {"GTC": <float>, "PBC": <float>, "PR": <float>}
\end{verbatim}
\end{tcolorbox}
\FloatBarrier

\subsubsection{Reproducibility and the Choice of a Proprietary VLM Judge}
\label{sec:supp_vlm_reproducibility}

Using Gemini~3.0~Flash~\cite{google2025gemini3} as the VLM-as-judge introduces a reproducibility limitation inherent to proprietary VLM models: the underlying model can change or be deprecated by its provider without notice, and its internals cannot be independently audited.
However, this limitation is not unique to our work: concurrent VTO evaluation work relies on proprietary VLM judges as well, with OpenVTON-Bench~\cite{li2026openvton} and CORAL~\cite{kim2026coral} using Gemini~2.0~Flash and GPT-5, respectively.
Open-weight VLM models currently lack the discriminative power that reliable VTO evaluation requires.
As an illustration, we repeat our GTC/PBC/PR evaluation with an open-weight VLM judge, \texttt{Qwen3-VL-8B-Instruct}~\cite{bai2025qwen3}, on the same five methods evaluated on VITON-HD-edit in {Tab.~\ref{tab:ctrl_quan}} of the main paper (the four proprietary editing models and \textbf{CtrlVTON}).
Table~\ref{tab:supp_qwen_judge} reports, for each method, the mean over five independent scoring runs together with the 95\% confidence interval.
The confidence intervals overlap substantially across methods on GTC and PR, and even where a gap appears (e.g., PBC) it spans only a few hundredths of a point---far too little to reflect the differences we know exist between these methods.
In practice, \texttt{Qwen3-VL-8B-Instruct} cannot reliably rank methods at this quality level.
Utilizing a proprietary model for VLM-as-judge is required to assess VTO output at retail-grade image quality.

\begin{table}[h]
\centering
\caption{VLM-as-judge scores from an open-weight model (Qwen3-VL-8B-Instruct~\cite{bai2025qwen3}) on VITON-HD-edit, using the same GTC/PBC/PR system prompt as Appendix~\ref{sec:supp_metrics_vlm}. Each cell reports the mean $\pm$ 95\% confidence interval over 5 independent runs. Compare against the proprietary-VLM (Gemini 3.0 Flash) scores for the same methods in {Tab.~\ref{tab:ctrl_quan}} of the main paper: the confidence intervals here overlap heavily despite the real quality differences that {Tab.~\ref{tab:ctrl_quan}} reveals.}
\label{tab:supp_qwen_judge}
\begin{tabular}{lccc}
\toprule
Method & GTC$\uparrow$ & PBC$\uparrow$ & PR$\uparrow$ \\
\midrule
Nano Banana Pro~\cite{google2025nanobananapro} & $4.782 \pm 0.031$ & $4.911 \pm 0.047$ & $4.902 \pm 0.041$ \\
GPT Image 1.5~\cite{openai2025gptimage}        & $4.803 \pm 0.044$ & $4.869 \pm 0.033$ & $4.921 \pm 0.041$ \\
Seedream 4.5~\cite{bytedance2025seedream4}     & $4.741 \pm 0.052$ & $4.933 \pm 0.021$ & $4.858 \pm 0.043$ \\
FLUX.2~[pro]~\cite{black2025flux2}             & $4.789 \pm 0.037$ & $4.902 \pm 0.041$ & $4.928 \pm 0.037$ \\
\midrule
\rowcolor{ourRow}
\textbf{CtrlVTON} & $4.790 \pm 0.018$ & $5.000 \pm 0.000$ & $4.970 \pm 0.009$ \\
\bottomrule
\end{tabular}
\end{table}

\subsection{Task-Following Consistency (TFC) for Token Control}
\label{sec:supp_tfc}
GTC/PBC/PR are designed to be task-agnostic: they measure properties of the output that should hold regardless of which task token ($\tau_{task}$) was used.
To quantitatively evaluate adherence to the task-token ({Sec.~\ref{sec:exp_vto}} of the main paper, ``Task-token adherence''), we additionally define a fourth, task-specific criterion used only for the evaluation in Tab.~\ref{tab:supp_task_token}.

\noindent\textbf{TFC} (Task-Following Consistency, $[0, 5]$): how faithfully the output realizes the operation specified by the task token, evaluated separately for each token:
\begin{itemize}
\item \textsc{full\_swap}: every garment of the matching class in the input is replaced with the reference; no original garment of that class remains.
\item \textsc{partial\_swap}: exactly one garment of the matching class is replaced with the reference; all other garments, including same-class items, remain unchanged.
\item \textsc{add}: all original garments are preserved; the reference is added as an additional layer on top of the existing outfit.
\end{itemize}
The grader is shown the input person, reference garment, generated output, and the intended task token, and returns a strict TFC score under the same protocol as Appendix~\ref{sec:supp_metrics_vlm}.
We are not aware of any open-weight VTO system that exposes operation-level control via discrete tokens, so a head-to-head comparison is not possible; Tab.~\ref{tab:supp_task_token} therefore reports CtrlVTON-base scores in isolation.
The takeaway is that the model achieves consistently high TFC across all three operations on OmniTry-Bench, indicating that the token interface is reliable, while GTC remains stable across tokens---token-level control does not trade off against garment fidelity.
The exact prompt we send to the model is below.

\begin{figure}[h]
\centering
\includegraphics[width=\linewidth]{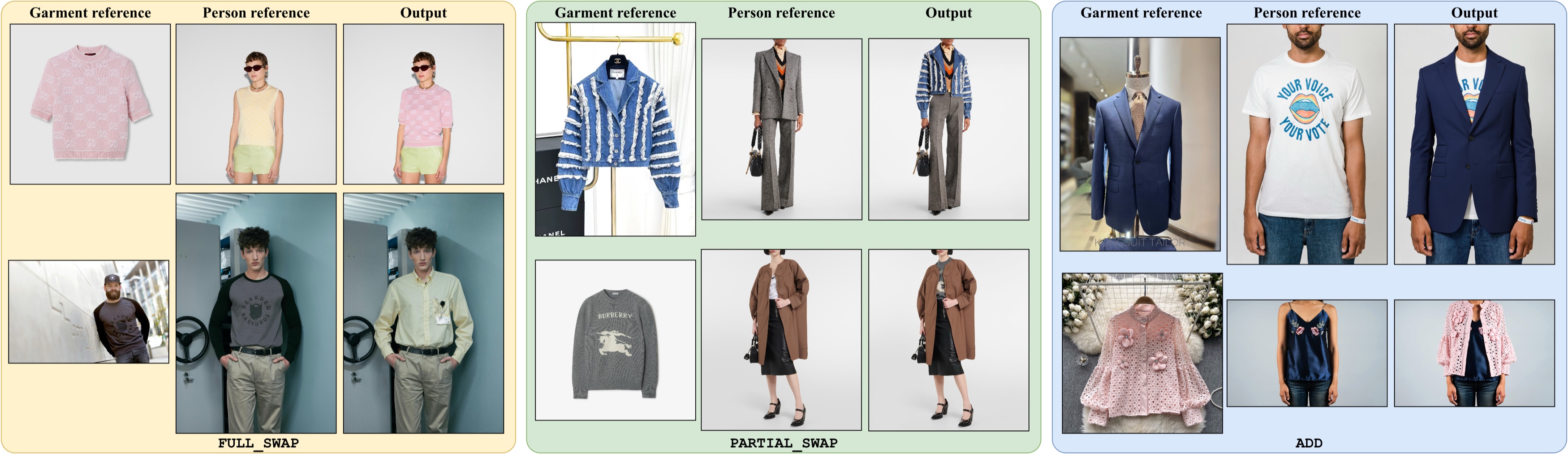}
\caption{\textbf{Qualitative results of task-token control on OmniTry-Bench.} The figure is organized into three panels, one per task token: \textbf{\textsc{full\_swap}} (left), \textbf{\textsc{partial\_swap}} (middle), and \textbf{\textsc{add}} (right). \textsc{full\_swap} replaces every garment of the matching class with the reference. \textsc{partial\_swap} replaces only the targeted item while leaving the rest of the outfit intact (e.g., the inner top is swapped while the outer jacket is preserved). \textsc{add} keeps every existing garment in place and layers the reference on top (e.g., a blazer layered over a tee).}
\label{fig:supp_task_token_qual}
\end{figure}

\begin{table}[h]
\centering
\caption{Task-token following on OmniTry-Bench (clothes-only subset). For each sample, we run \textbf{CtrlVTON-base} under the three task tokens, holding all other inputs fixed; only the task token varies between rows. \textbf{TFC} is the task-specific metric defined above; \textbf{GTC} is the same garment-fidelity metric as in {Tab.~\ref{tab:single_quan}} of the main paper. Scores are on $[0, 5]$ scale, higher is better.}
\label{tab:supp_task_token}
\setlength{\tabcolsep}{8pt}
\begin{tabular}{lcc}
\toprule
Task token & TFC$\uparrow$ & GTC$\uparrow$ \\
\midrule
\textsc{full\_swap}    & 4.9234 & 4.1401 \\
\textsc{partial\_swap} & 4.5872 & 4.1356 \\
\textsc{add}           & 4.7156 & 4.1287 \\
\midrule
Average                & 4.7421 & 4.1348 \\
\bottomrule
\end{tabular}
\end{table}

\FloatBarrier
\begin{tcolorbox}[
  colback=gray!5, colframe=gray!50!black,
  title={VLM-as-judge prompt for TFC (verbatim)},
  fonttitle=\bfseries\small,
  fontupper=\scriptsize,
  boxrule=0.4pt, arc=1pt, left=4pt, right=4pt, top=4pt, bottom=4pt,
  breakable
]
\begin{verbatim}
[Same preamble and image definitions as the GTC/PBC/PR prompt.]

You will also be given a TASK TOKEN that specifies the intended
operation:
  [TASK_TOKEN]        - one of: full_swap, partial_swap, add.

Score the GENERATED_IMAGE on the following axis. The score is a
float in [0, 5]. Be STRICT: 5 is reserved for outputs with no
perceivable issues under detailed inspection; deduct points for any
deviation from the intended operation, however small.

  TFC (Task-Following Consistency, [0, 5]):
    Does the output faithfully realize the operation specified by
    the TASK_TOKEN?

    full_swap:
      - Every garment of the matching class in INPUT_PERSON is
        replaced with the REFERENCE_GARMENT.
      - No original garment of that class remains.

    partial_swap:
      - Exactly one garment of the matching class is replaced with
        the REFERENCE_GARMENT.
      - All other garments, including other same-class items, remain
        unchanged.

    add:
      - All original garments in INPUT_PERSON are preserved.
      - The REFERENCE_GARMENT is added as an additional layer on top
        of the existing outfit.

Return your answer as a single JSON object, no extra text:
  {"TFC": <float>}
\end{verbatim}
\end{tcolorbox}
\FloatBarrier

\subsection{Mask-Adherence Metrics (IoU, $d_{\text{Hu}}$, $d_H$)}
\label{sec:supp_mask_metrics}

For each generated try-on image we re-extract the garment mask $M_{\text{gen}}$ via VIP-SAM ({Sec.~\ref{sec:vipseg}} of the main paper) and compare it against the input control mask $M_p$ under three complementary criteria---region overlap, global shape, and boundary deviation (illustrated in Fig.~\ref{fig:supp_mask_metric_visual}).

\paragraph{Intersection over Union (IoU).}
IoU measures the overlap between the control mask and the generated garment region:
\begin{equation}
\mathrm{IoU}(M_p, M_{\text{gen}}) = \frac{|M_p \cap M_{\text{gen}}|}{|M_p \cup M_{\text{gen}}|}.
\end{equation}
Higher is better.
IoU captures region-level agreement but is insensitive to localized boundary failures and shape distortions that preserve overall area.

\paragraph{Hu moment distance $d_{\text{Hu}}$.}
To capture \emph{global shape} agreement beyond mere overlap, we adopt the definition of Hu-moment from FitControler~\cite{yang2025fitcontroler}.
For a binary mask $I(x,y)$, the central moments are
\begin{equation}
\mu_{pq} = \sum_{x,y} (x-\bar{x})^p (y-\bar{y})^q \, I(x,y),
\end{equation}
where $(\bar{x}, \bar{y})$ is the centroid.
The moments are normalized by area as
\begin{equation}
\eta_{pq} = \frac{\mu_{pq}}{\mu_{00}^{\,r}}, \qquad r = \tfrac{p+q}{2} + 1,
\end{equation}
from which the seven translation-, scale-, and rotation-invariant Hu invariants $\Phi(I) = [\phi_1, \ldots, \phi_7]$~\cite{hu1962moments} can be obtained (\eg, $\phi_1=\eta_{20}+\eta_{02}$).
We then report the Euclidean distance between the invariant vectors of the control mask and the re-extracted mask:
\begin{equation}
d_{\text{Hu}}(M_p, M_{\text{gen}}) = \bigl\lVert \Phi(M_p) - \Phi(M_{\text{gen}}) \bigr\rVert_2.
\end{equation}
Lower is better.

\paragraph{Symmetric Hausdorff distance $d_H$.}
Hu moments are useful for describing global shape but are insensitive to \emph{localized} failures such as a sleeve protruding well beyond the requested mask.
Following~\cite{huttenlocher1993hausdorff,yang2025fitcontroler}, we additionally report the symmetric Hausdorff distance between contours defined by point sets $A = \partial M_p$ and $B = \partial M_{\text{gen}}$:
\begin{equation}
d_H(A, B) = \max\!\left\{ \sup_{a \in A} \inf_{b \in B} d(a,b),\;\; \sup_{b \in B} \inf_{a \in A} d(b,a) \right\},
\end{equation}
with $d(\cdot,\cdot)$ the Euclidean distance in pixels.
Lower is better.
Together with IoU and $d_{\text{Hu}}$, this gives a balanced view of region, global shape, and worst-case boundary fidelity.

\subsection{Supplementary evaluation of mask-adherence metrics using SAM3}
\label{sec:supp_self_ref}

Both the garment-fidelity metrics (Appendix~\ref{sec:supp_metrics_object}) and the mask-adherence metrics above rely on VIP-SAM to extract garment crops and re-extract $M_{\text{gen}}$ from generated images.
Because VIP-SAM is also used throughout the training-data pipeline ({Sec.~\ref{sec:cvton_data}} of the main paper) to construct $M_p$ and $M_{g_{\text{ref}}}$, evaluating with VIP-SAM raises a concern of \emph{self-referential evaluation}: strong scores could in principle reflect that CtrlVTON has overfit to VIP-SAM's particular notion of a garment mask, or that some information leaks from training into the evaluation metric through the shared segmenter, rather than reflecting genuine spatial controllability.

To rule this out, we re-run inference-time mask extraction and metric computation on VITON-HD-edit using an independent segmenter, SAM3~\cite{carion2025sam3}, prompted with a text description of the garment rather than a visual prompt---so that neither the segmenter nor the prompting modality overlaps with anything used at training time.
Table~\ref{tab:supp_sam3_reeval} reports the resulting scores for all five methods on VITON-HD-edit.
Compared to the VIP-SAM-based scores in {Tab.~\ref{tab:ctrl_quan}} of the main paper, all five methods shift only slightly under SAM3, but critically, the ranking is unchanged: CtrlVTON remains far ahead of every proprietary baseline on all three spatial-control metrics, so the conclusion of {Tab.~\ref{tab:ctrl_quan}} of the main paper is unaffected by the choice of segmenter.
We also observed a small number of cases where SAM3 extracts a garment mask that differs noticeably from the corresponding VIP-SAM mask; Fig.~\ref{fig:supp_sam3_maskdiff} shows that in these cases, CtrlVTON's output still closely follows whichever mask it is given.
Taken together, the table and the figure indicate that CtrlVTON's spatial controllability is not an artifact of overfitting to VIP-SAM's specific masks, nor of train--test leakage through the evaluation metric itself.

\begin{table}[h]
\centering
\caption{Re-evaluation of single-garment mask-controllable VTO on VITON-HD-edit using SAM3 (text-prompted) instead of VIP-SAM for both inference-time mask extraction and metric computation. Compare against {Tab.~\ref{tab:ctrl_quan}} of the main paper, which uses VIP-SAM throughout.}
\label{tab:supp_sam3_reeval}
\small
\setlength{\tabcolsep}{5pt}
\begin{tabular}{lccccc}
\toprule
Method & IoU$\uparrow$ & $d_{\text{Hu}}\downarrow$ & $d_H\downarrow$ & M-DINO$\uparrow$ & M-CLIP-I$\uparrow$ \\
\midrule
Nano Banana Pro~\cite{google2025nanobananapro} & 0.858 & 0.0049 & 37.12 & 0.8201 & 0.9043 \\
GPT Image 1.5~\cite{openai2025gptimage}        & 0.798 & 0.0081 & 55.67 & 0.7798 & 0.8829 \\
Seedream 4.5~\cite{bytedance2025seedream4}     & 0.852 & 0.0043 & 43.87 & 0.8102 & 0.9008 \\
FLUX.2~[pro]~\cite{black2025flux2}             & 0.861 & 0.0058 & 40.15 & 0.7889 & 0.8981 \\
\midrule
\rowcolor{ourRow}
\textbf{CtrlVTON} & 0.9589 & 0.0035 & 22.69 & 0.8193 & 0.8941 \\
\bottomrule
\end{tabular}
\end{table}

\begin{figure}[h]
\centering
\includegraphics[width=\linewidth]{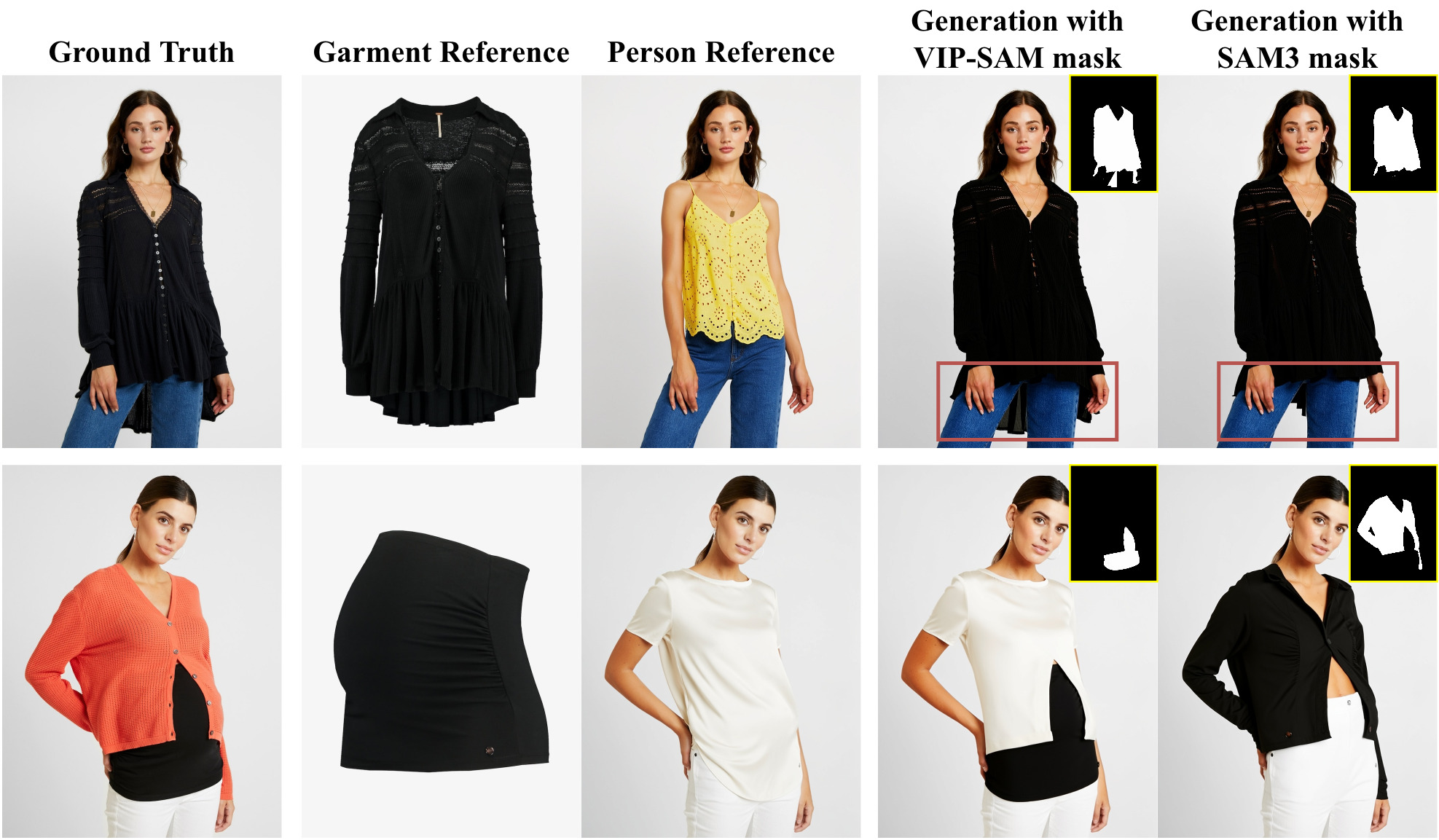}
\caption{\textbf{CtrlVTON generalizes across segmentation models.} These examples illustrate the cases where VIP-SAM and SAM3 disagree on the garment mask. For each example (row), we extract the garment mask from the ground-truth image using two different segmentation models---VIP-SAM and SAM3, shown inset with a yellow border in the last two columns---and feed each mask to CtrlVTON alongside the same garment and person references.
\textbf{Row~1:} the two masks disagree on the garment's hemline length; CtrlVTON follows each mask faithfully, generating a correspondingly shorter or longer hem rather than a fixed length learned from training.
\textbf{Row~2:} VIP-SAM's mask captures the \emph{garment reference} correctly, while SAM3's text-prompted mask captures the cardigan; CtrlVTON again follows whichever mask it receives, producing markedly different outputs.
In both cases, the generated output tracks the given mask rather than a shape prior learned from VIP-SAM's masks during training, showing that CtrlVTON's spatial controllability transfers to masks from a segmentation model it never saw at training time.}
\label{fig:supp_sam3_maskdiff}
\end{figure}

\clearpage
\section{Additional Qualitative Results}
\label{sec:supp_qualitative}

This section collects qualitative comparisons that did not fit in the main paper.
Each figure compares along a specific axis; the caption gives the full context.

\begin{figure}[h]
\centering
\includegraphics[width=\linewidth]{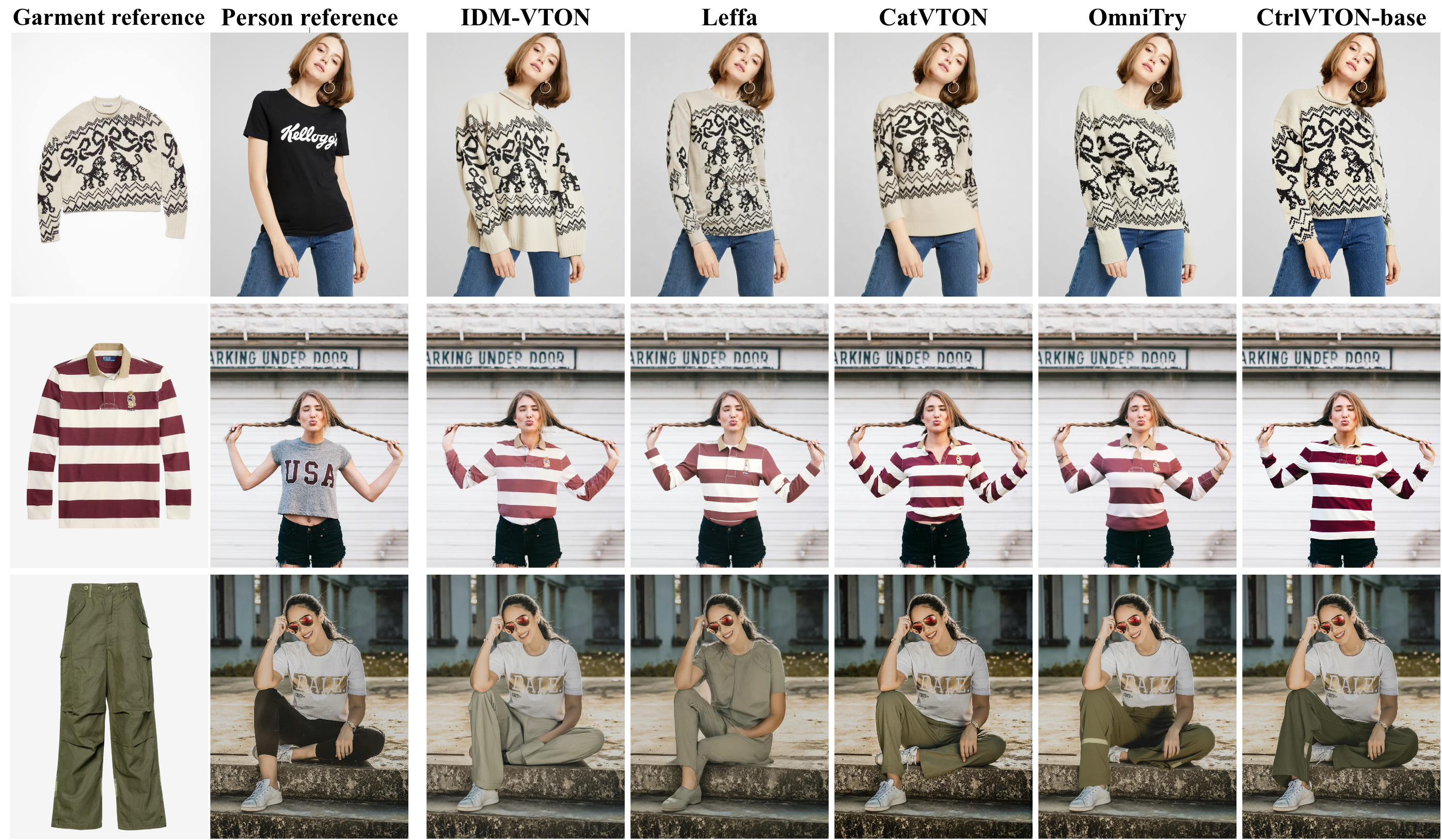}
\caption{\textbf{Additional single-garment VTO results.} Comparisons across diverse garment types, person images, and reference representations, complementing Fig.~\ref{fig:vto_qual}.}
\label{fig:supp_qual_single}
\end{figure}

\begin{figure}[h]
\centering
\includegraphics[width=\linewidth]{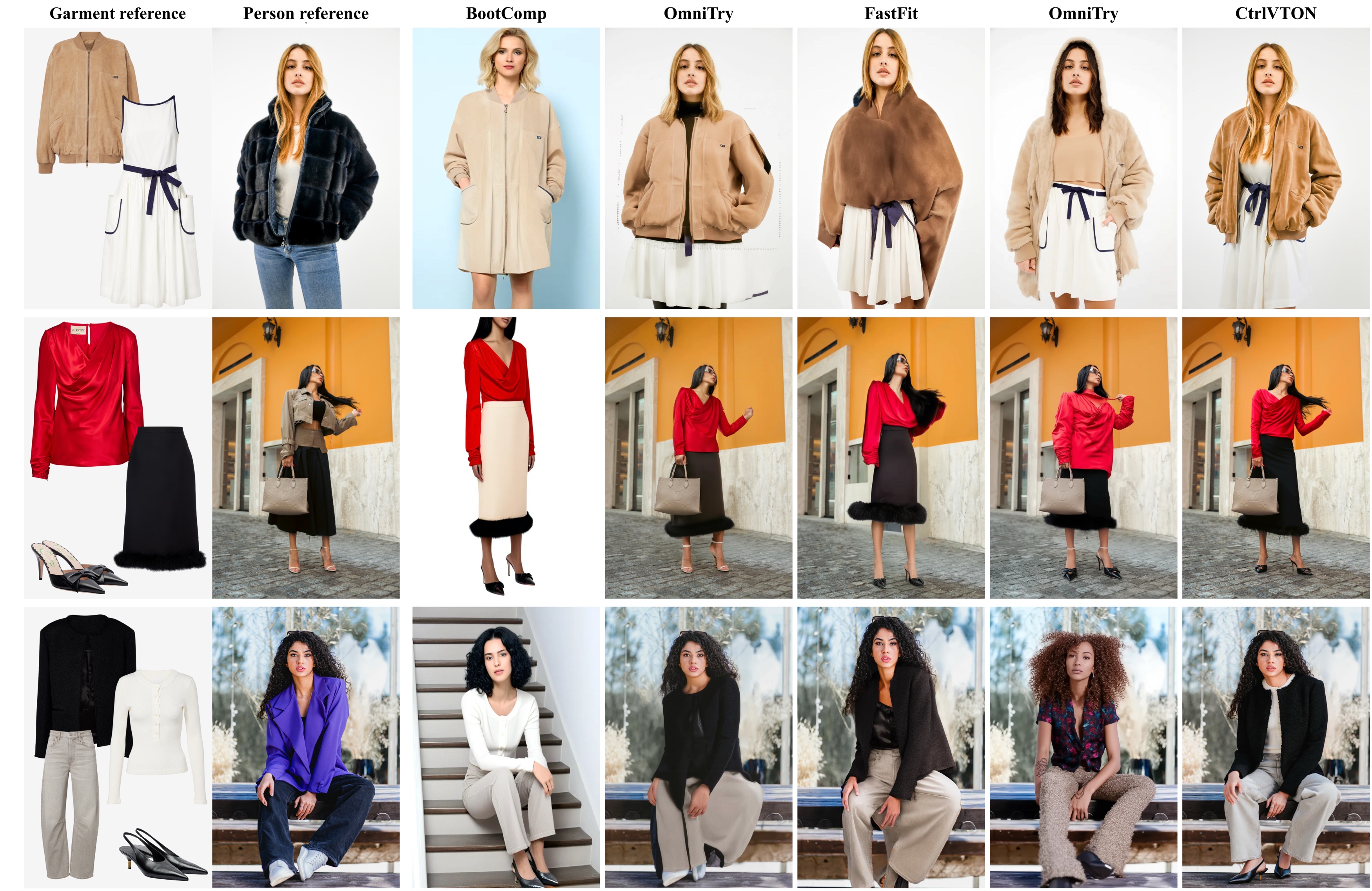}
\caption{\textbf{Additional multi-garment VTO results.} Each row shows multiple reference garments tried on a single person, comparing CtrlVTON-base (which composes all garments in a single forward pass) with multi-garment baselines. FastFit and BootComp natively accept multiple garments, while FitDit and OmniTry are run sequentially, applying one garment at a time.}
\label{fig:supp_qual_multi}
\end{figure}

\begin{figure}[h]
\centering
\includegraphics[width=\linewidth]{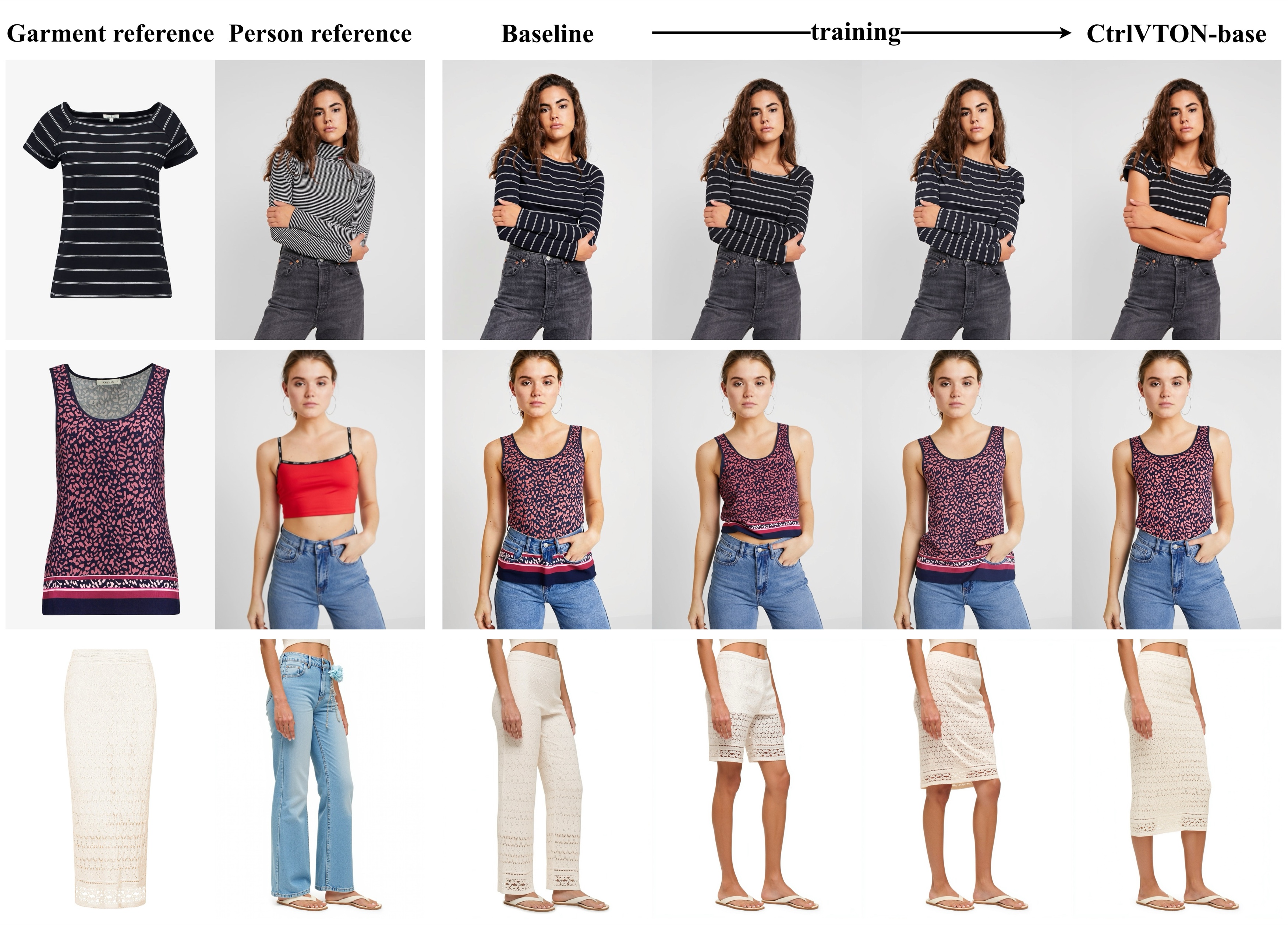}
\caption{\textbf{Effect of data curation on try-on quality.} Comparison between a baseline (left) and our \textbf{CtrlVTON-base} trained on curated data (right). The baseline exhibits the silhouette-leakage artifact described in Appendix~\ref{sec:supp_qc}: the synthesized garment's outline follows the silhouette of the garment worn by the reference person, rather than that of the reference garment (rows 1, 3). It also produces physically implausible results, such as unnatural garment fit and body contact (row 2). CtrlVTON-base, trained on triplets that pass our three-stage quality-control funnel, resolves both issues and renders $g_{\text{ref}}$ faithfully. This demonstrates that a well-curated supervision set is sufficient to turn the same backbone into a high-quality VTO model.}
\label{fig:supp_qual_silhouette}
\end{figure}

\begin{figure}[h]
\centering
\includegraphics[width=\linewidth]{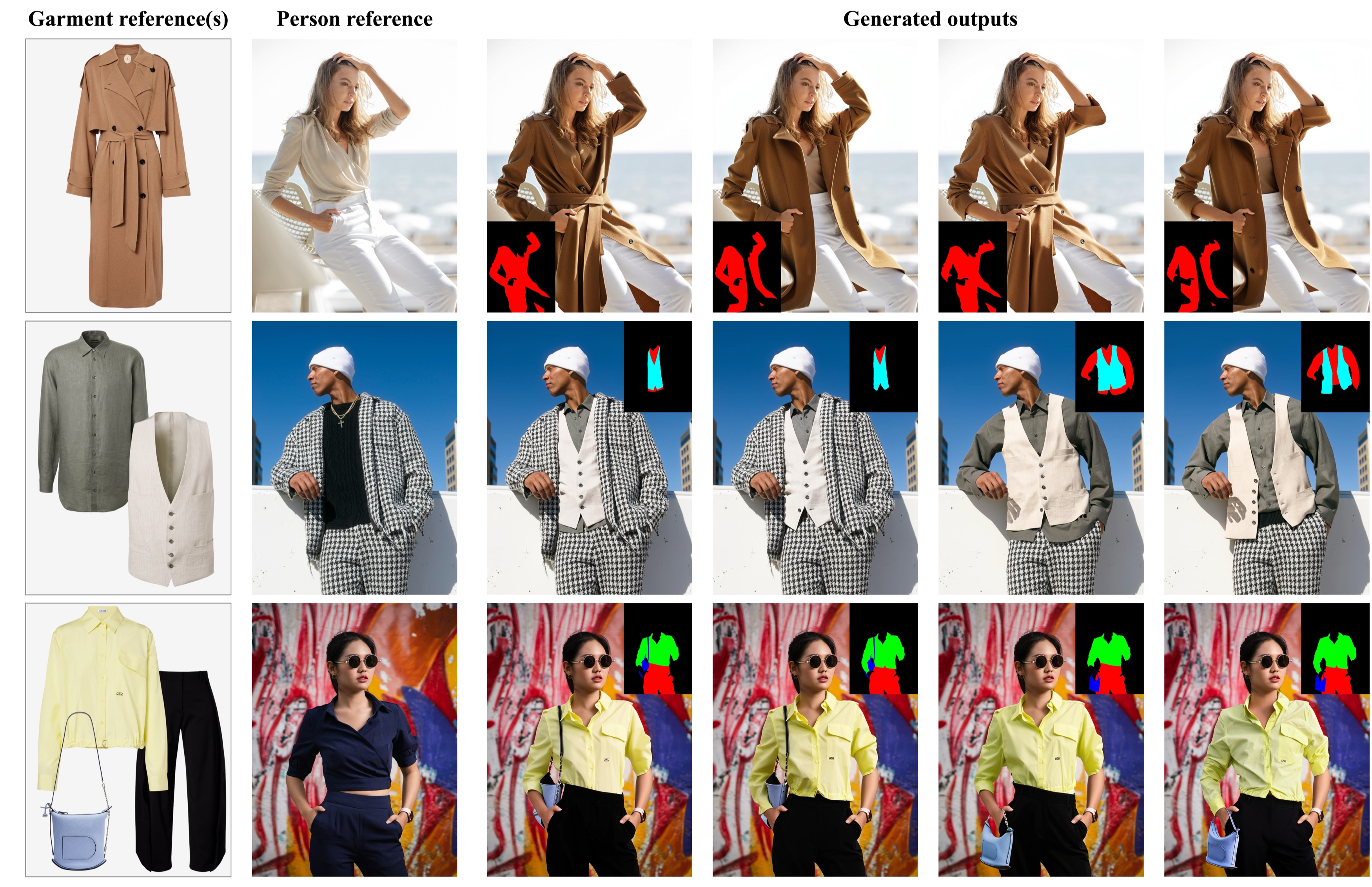}
\caption{\textbf{Additional mask-controllable try-on results.} Each row shows multiple outputs from the same (person, garment) pair generated by varying only the input control mask, complementing {Fig.~\ref{fig:ctrlvto_qual}} of the main paper.}
\label{fig:supp_qual_ctrl}
\end{figure}

\begin{figure}[h]
\centering
\includegraphics[width=\linewidth]{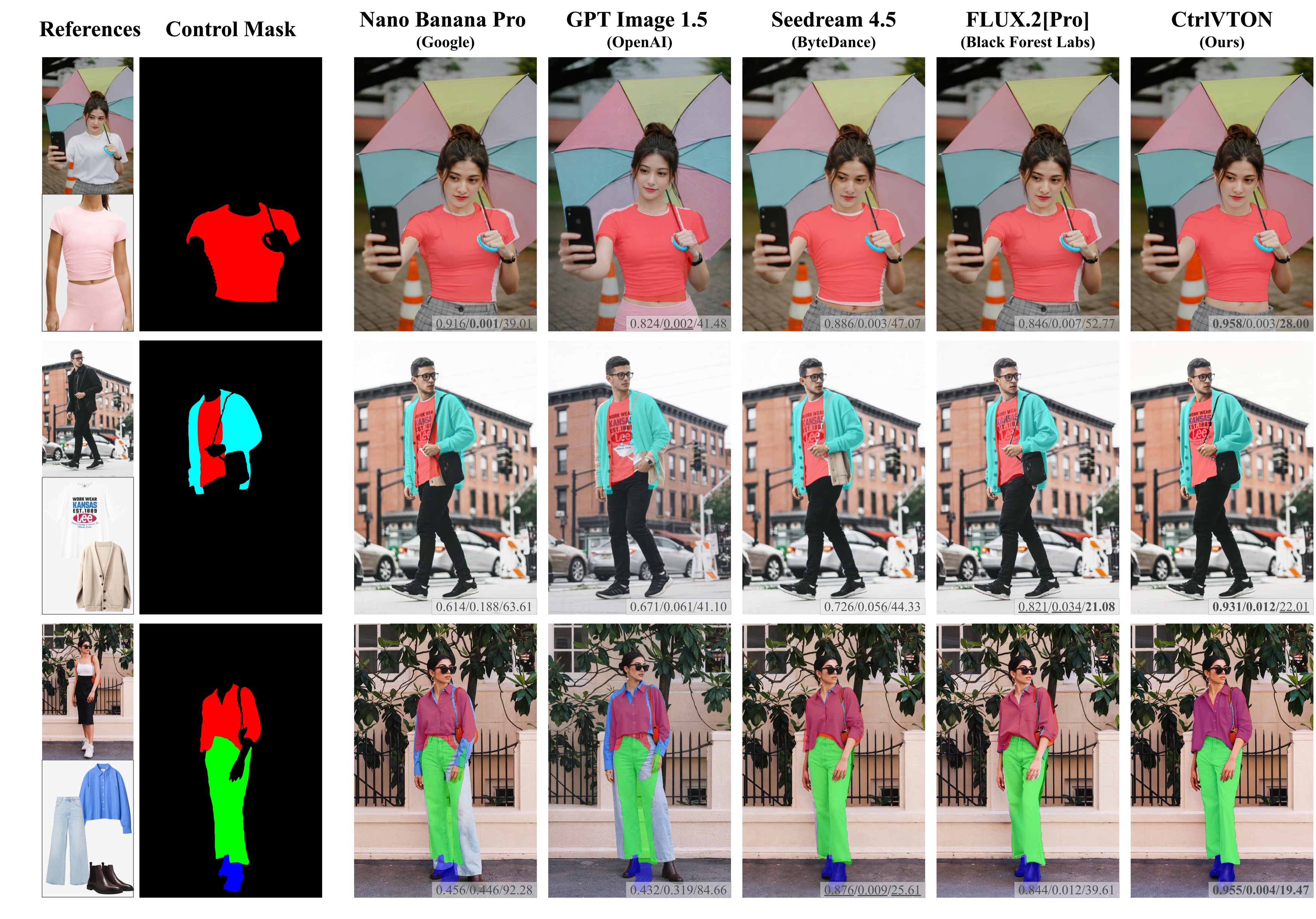}
\caption{\textbf{Visualization of mask-following metrics.} Each output is overlaid with the input control mask $M_p$ and the re-extracted mask $M_{\text{gen}}$ (obtained via VIP-SAM), with per-image IoU, $d_{\text{Hu}}$, and $d_H$ printed alongside. Visually large discrepancies translate into low IoU and high $d_H$, while subtle global-shape mismatches are reflected in $d_{\text{Hu}}$. The overlays confirm that CtrlVTON respects the requested layout while proprietary baselines don't.}
\label{fig:supp_mask_metric_visual}
\end{figure}

\begin{figure}[h]
\centering
\includegraphics[width=\linewidth]{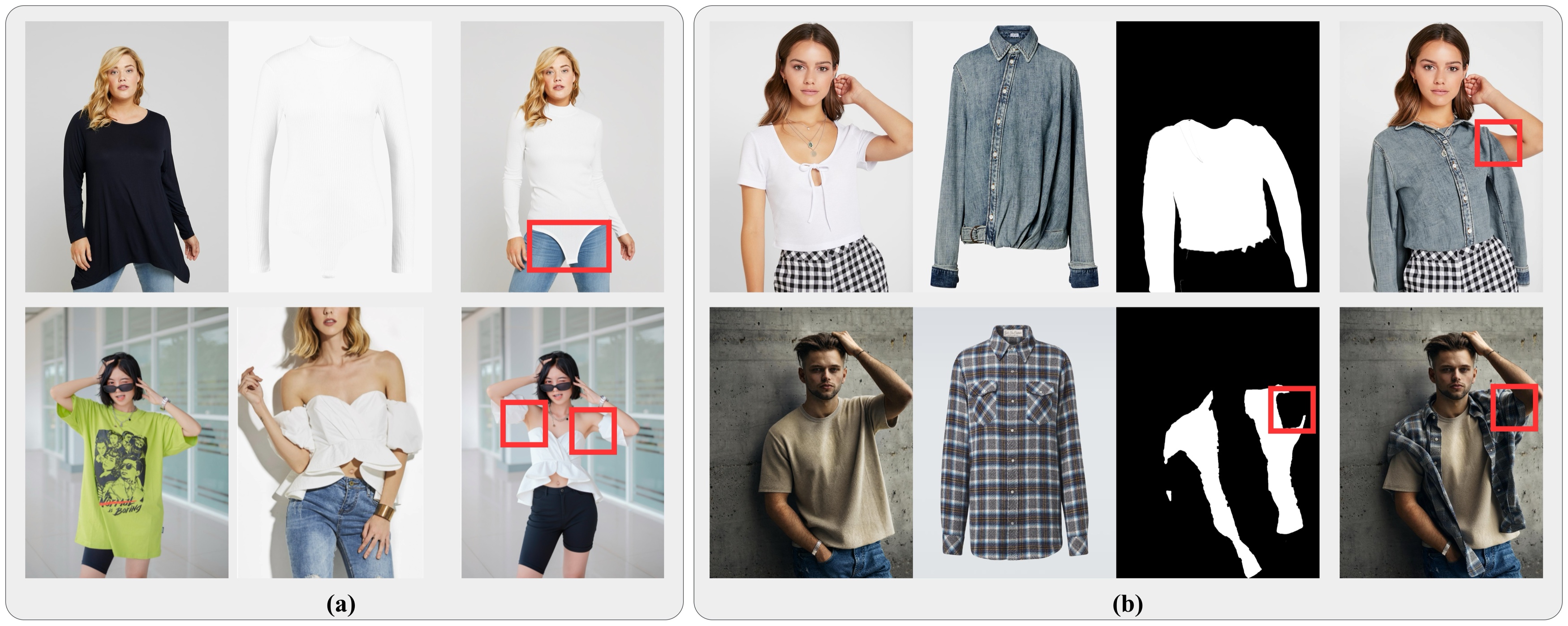}
\caption{\textbf{Representative failure cases of our two models.}
\textbf{(a) CtrlVTON-base} is guided only by the semantic tokens (garment-class, task), so it can render the reference garment in a stylistically or physically implausible way, e.g., a one-piece bodysuit left untucked over pants, or a top whose fabric appears torn at the torso as if split by the body.
\textbf{(b) CtrlVTON} is highly responsive to the user-provided control mask, which is also its main weakness. A poorly drawn mask can cause the model to ignore the person reference and synthesize the garment in arbitrary regions, or conversely to extend the garment beyond the masked area. Output quality therefore depends directly on the quality of the input mask.}
\label{fig:supp_qual_limitations}
\end{figure}

\end{document}